\documentclass[letterpaper, 10 pt, conference]{ieeeconf}  

\IEEEoverridecommandlockouts                              

\usepackage{graphics} 
\usepackage{graphicx}
\usepackage{epsfig} 
\usepackage{times} 
\usepackage{amsmath} 
\usepackage{amssymb}  

\usepackage{bm}
\usepackage[linesnumbered,ruled,vlined]{algorithm2e}
\usepackage[noend]{algpseudocode}
\usepackage[font={small}]{caption}
\usepackage{subcaption}
\usepackage{physics}
\usepackage{xcolor}
\usepackage{tikz}
\usetikzlibrary {arrows.meta}
\usetikzlibrary{fit,
                positioning}
\usepackage[belowskip=1ex]{subcaption}  

\usetikzlibrary{calc} 
\usetikzlibrary{positioning}
\usepackage{balance}
\usepackage{soul}
\usetikzlibrary{arrows}
\usepackage{cite}
\usepackage{url}
\usepackage{multirow}
\usepackage{booktabs} 

\usepackage{amssymb}
\usepackage{amsfonts}
\usepackage{amsmath}
\usepackage{amsthm}
\usepackage{bm}

\usepackage[normalem]{ulem}                                        
\usepackage{marginnote}
\usepackage[textwidth=10ex,colorinlistoftodos]{todonotes}

\colorlet{mh}{red}
\colorlet{fwu}{red}
\colorlet{ywu}{blue}
\colorlet{kchen}{blue}
\colorlet{lchen}{green}
\colorlet{zbing}{green}
\colorlet{shaddadin}{purple}
\colorlet{iperez}{cyan}
\colorlet{schneider}{magenta}

\usepackage{amssymb}
\usepackage{mathtools}
\usepackage{sansmath}

\mathchardef\mhyphen="2D   

\newcommand{\RNum}[1]{\uppercase\expandafter{\romannumeral #1\relax}}

\usepackage{setspace}
\title{\LARGE \bf

Flexible Informed Trees (FIT*): Adaptive Batch-Size Approach in Informed Sampling-Based Path Planning 
}

\author{Liding Zhang$^{1}$, Zhenshan Bing$^{1}$, Kejia Chen$^{1}$, Lingyun Chen$^{1}$, Kuanqi Cai$^{1}$, Yu Zhang$^{1}$, Fan Wu$^{1}$,\\ Peter Krumbholz$^{2}$, Zhilin Yuan$^{2}$, Sami Haddadin$^{1}$,  Alois Knoll$^{1}$ 
\thanks{$^{1}$L. Zhang, Z. Bing, K. Chen, L. Chen, K. Cai, Y. Zhang, F. Wu,  S. Haddadin and A. Knoll are with the Department of Informatics, Technical University of Munich, Germany.
{\tt\small liding.zhang@tum.de}}%
\thanks{$^{2}$P. Krumbholz, Z. Yuan are with the Department of Technology \& Innovation, KION Group Linde Material Handling GmbH, Germany.}%
}
\begin{document}

\maketitle
\thispagestyle{empty}
\pagestyle{empty}

\begin{abstract}
In path planning, anytime almost-surely asymptotically optimal planners dominate the benchmark of sampling-based planners. A notable example is Batch Informed Trees (BIT*), where planners iteratively determine paths to batches of vertices within the exploration area. However, utilizing a consistent batch size is inefficient for initial pathfinding and optimal performance, it relies on effective task allocation.
This paper introduces Flexible Informed Trees (FIT*), a sampling-based planner that integrates an adaptive batch-size method to enhance the initial path convergence rate. FIT* employs a flexible approach in adjusting batch sizes dynamically based on the inherent dimension of the configuration spaces and the hypervolume of the $n$-dimensional hyperellipsoid. 
By applying dense and sparse sampling strategy, FIT* improves convergence rate while finding successful solutions faster with lower initial solution cost.
This method enhances the planner's ability to handle confined, narrow spaces in the initial finding phase and increases batch vertices sampling frequency in the optimization phase. FIT* outperforms existing single-query, sampling-based planners on the tested problems in $\mathbb{R}^2$ to $\mathbb{R}^8$, and was demonstrated on a real-world mobile manipulation task.


\end{abstract}

\section{Introduction}

Path planning is essential in robotics, autonomous driving, game design, and related fields. Early methods like Dijkstra's algorithm~\cite{dijkstra1959note} found the shortest path in a graph, while A*\cite{hart1968formal} integrated graph and heuristic search for faster pathfinding. Anytime Repairing A* (ARA*)\cite{likhachev2003ara} offers progressive solutions during execution, catering to real-time scenarios with its \textit{Anytime} feature. Its \textit{Repairing} aspect also refines solutions by adjusting search parameters and balancing speed and quality.

Using graph-based planning in continuous state spaces is challenging due to the need for suitable discretization, known as \textit{prior discretization}. Coarse resolution boosts efficiency but may yield suboptimal paths. Finer resolution improves path quality~\cite{bertsekas1975convergence} but demands exponentially more computation, especially in high-dimensional environments, which are known as \textit{the curse of dimensionality}~\cite{bellman1957dynamic}.
To tackle this issue, sampling-based planning methods like Rapidly-exploring Random Trees (RRT)\cite{lavalle2001randomized} and Probabilistic Roadmaps (PRM)\cite{kavraki1996probabilistic} construct feasible paths by randomly sampling vertices in the \textit{configuration space} (\textit{$\mathcal{C}$-space}) and connects samples using local planners with collision checks. These approaches are suitable for high-dimensional and narrow environments.

\begin{figure}[t]
    \centering
    \begin{tikzpicture}

    \node[inner sep=0pt] (russell) at (-2.5,0)
    {\includegraphics[width=0.47\textwidth]{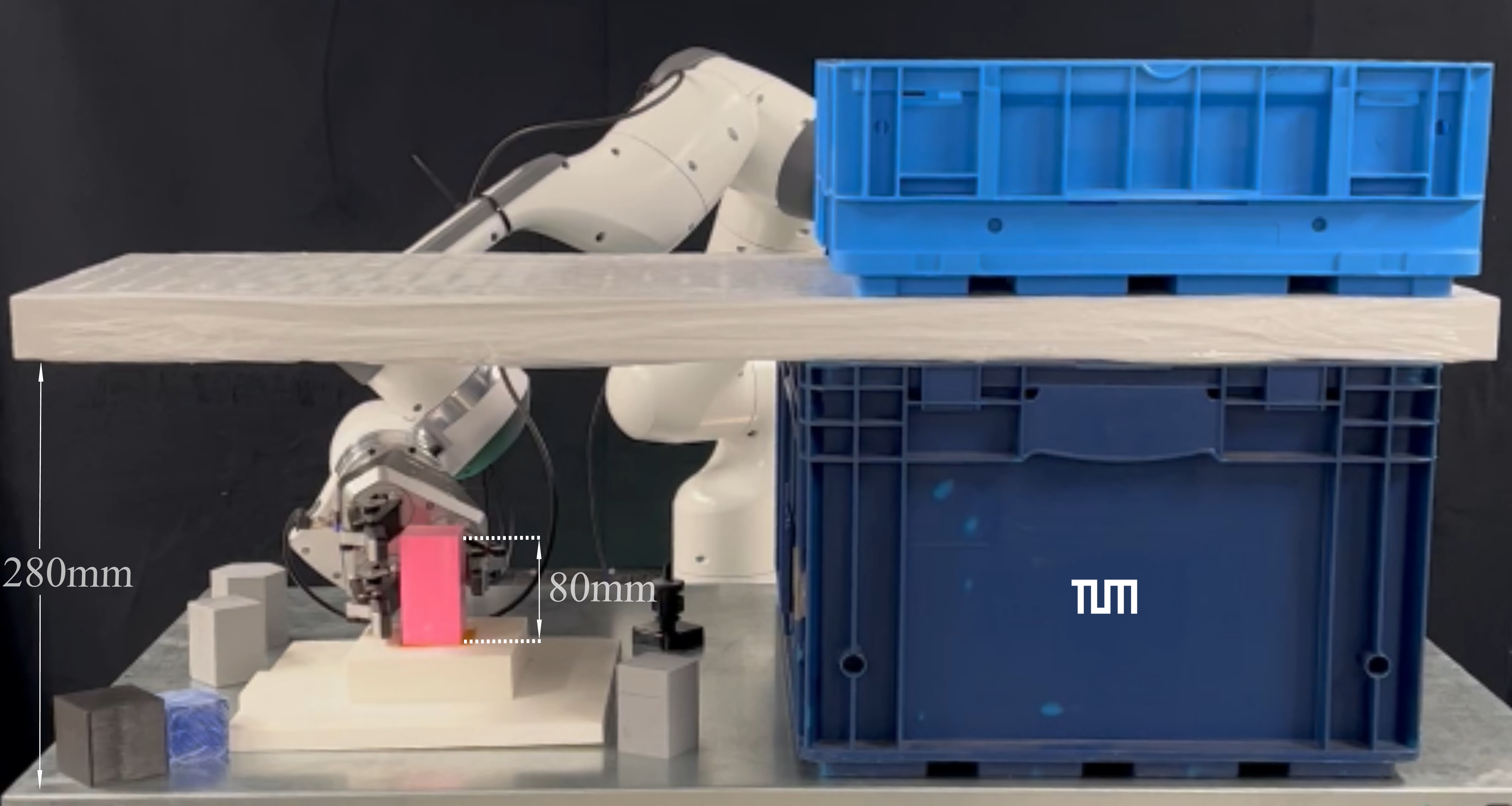}};

    \end{tikzpicture}
    \caption{FIT* on mobile manipulator robot during a real-time pull-out and place task in supermarket cell (Section~\ref{subsec:realExpri}).}
    \label{fig: darko_setup}
    \vspace{-1.7em} 
\end{figure}

Efficient pathfinding with high-dimensional \textit{$\mathcal{C}$-space} in an incremental planner hinges on balancing exploration and exploitation.
Batch Informed Trees (BIT*)\cite{gammell2015batch, gammell2020batch} compactly groups states into an implicit \textit{random geometric graph} (RGG)\cite{penrose2003random}, employing step-wise search akin to Lifelong Planning A* (LPA*)\cite{koenig2004lifelong} based on expected solution quality. Advanced BIT* (ABIT*)\cite{strub2020advanced} uses inflation and truncation factors to balance the exploration and exploitation of an increasingly dense RGG approximation. Adaptively Informed Trees (AIT*)\cite{strub2022adaptively,strub2020adaptively} utilized an asymmetrical search method with sparse collision checks in the reverse search. 
Effort Informed Trees (EIT*)\cite{strub2022adaptively}, a state-of-the-art planner uses admissible cost (i.e., lower bound on the true value) and effort heuristics to enhance its capability to tackle intricate scenarios, particularly in objectives with obstacle clearance, which is not estimable with the Euclidean distance heuristic. However, these planners lack adaptability in selecting an optimal batch size during the planning process. Optimizing batch size is advisable, considering variations among robots, scenarios, dimensions, and state spaces. Fewer samples might lower the probability of sampling vertices in narrow corridors. After the informed set (Fig.~\ref{fig: elipse}) contracts, more samples might lead to a more time-consuming edge check. This limitation hampers overall planning efficiency\cite{gammell2020batch}.

This paper presents Flexible Informed Trees (FIT*), which integrates the strengths of both \textit{search strategies} and \textit{approximation techniques}. FIT* aims to integrate adaptive batch-size features to tackle wall gap challenges and efficient planning.
By employing decay-based sigmoid functions to leverage the decay factor for dynamic batch size adjustments, a smoothing operation is integrated to prevent large differences in resizing steps while maintaining optimality. This integration enhances the efficiency of the approximation process.

The practical efficacy of FIT* has been thoroughly demonstrated through real-world applications, as depicted in Fig.~\ref{fig: darko_setup} and~\ref{fig:simulation}. We evaluated autonomous mobile manipulator navigation performance. Its adaptability in dynamically adjusting batch sizes during optimization demonstrated notable enhancements in computational time efficiency and solution quality. FIT* addresses the limitations of existing informed heuristic sampling-based  algorithms by computing appropriate batch size during path optimization. FIT* proposed a potential research direction, especially in dealing with densely populated settings. The effectiveness of FIT* underlines its potential in autonomous robotics and path planning.

The contributions of this work are summarized as follows:
\begin{itemize}
    \item \textit{Efficient initial pathfinding}: FIT*'s dense sampling strategy in the initial pathfinding phase tackles difficult-to-sample regions problem, reducing the initial pathfinding time in various test environments up to approx. 24\%.
    \item \textit{Batch sampling frequency}: The sparse sampling strategy in the path-optimized phase benefits the search process by frequently sampling points in the informed set. Leads to a higher probability of sampling \textit{key} points.
    \item \textit{Adaptability to confined environments}: FIT* was tested in real-world applications, including mobile manipulator pull-out and place tasks in narrow spaces. FIT* showed a 27.78\% solution cost reduction in experiments.
\end{itemize}

\section{Related Work}

\begin{figure}[t!]
    \centering
    \begin{tikzpicture}
    \node[inner sep=0pt] (russell) at (-3,0)
    {\includegraphics[width=0.24\textwidth]{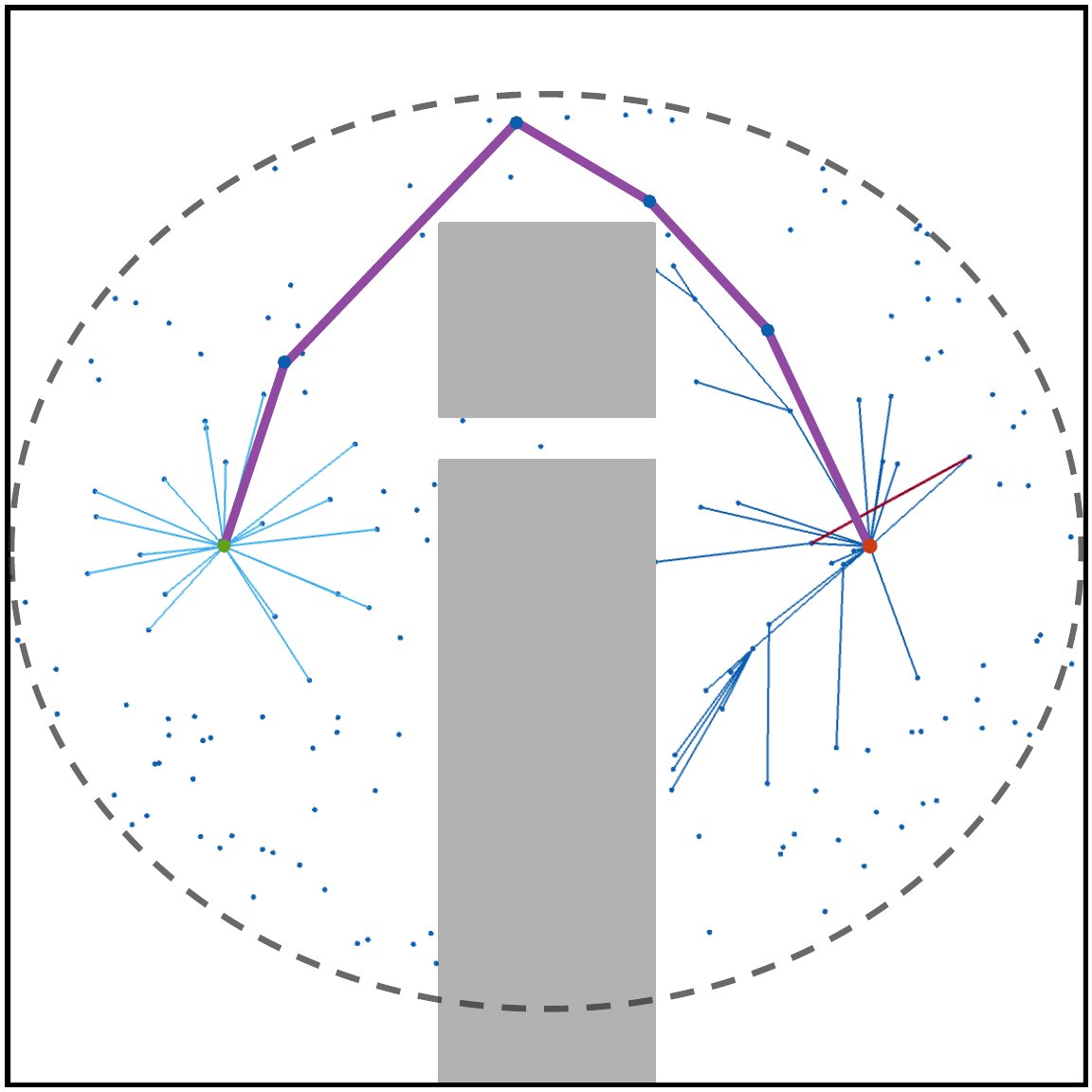}};
    \node[inner sep=0pt] (russell) at (1.4,0)
    {\includegraphics[width=0.24\textwidth]{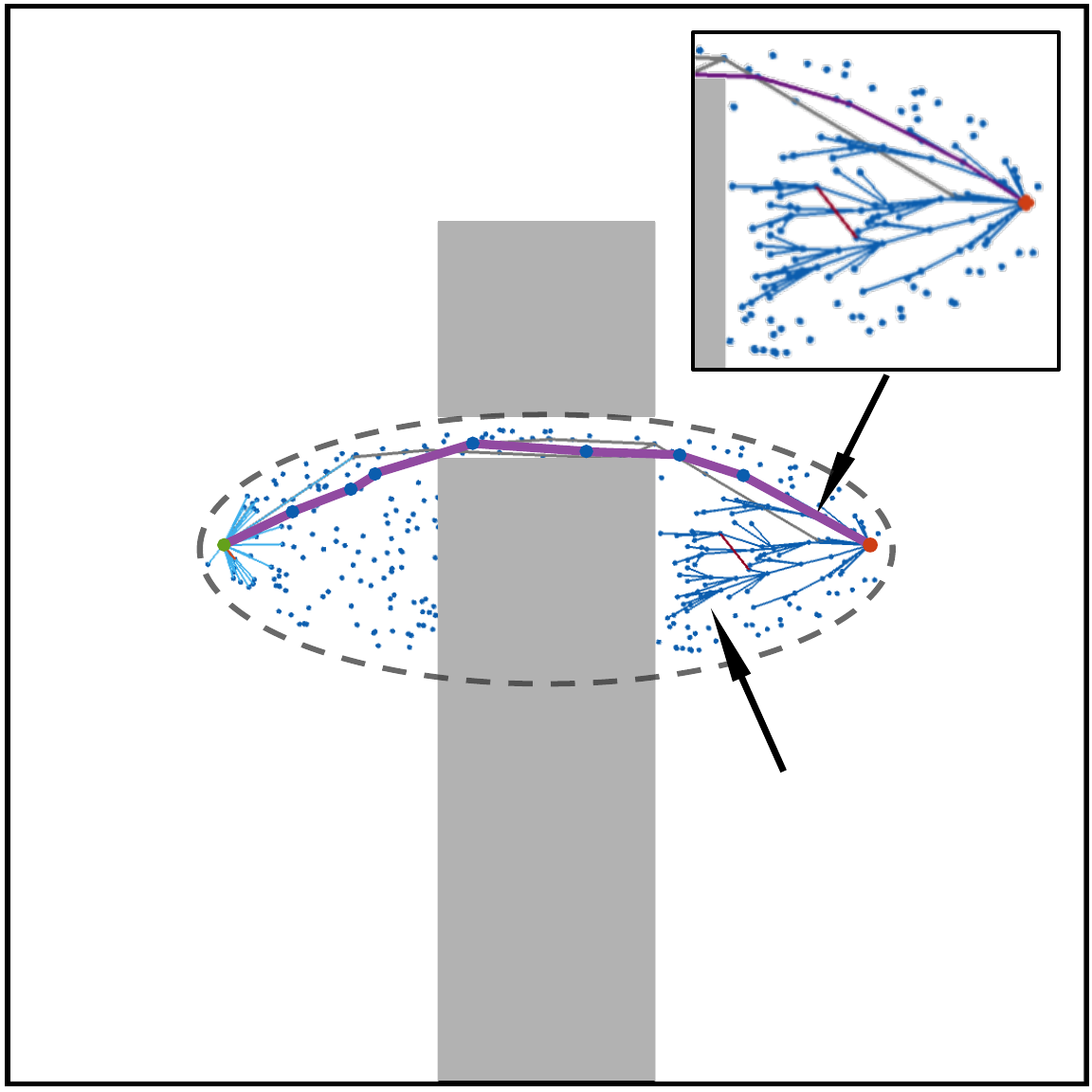}};
    \node[inner sep=0pt] (russell) at (-3,-5.2)
    {\includegraphics[width=0.24\textwidth]{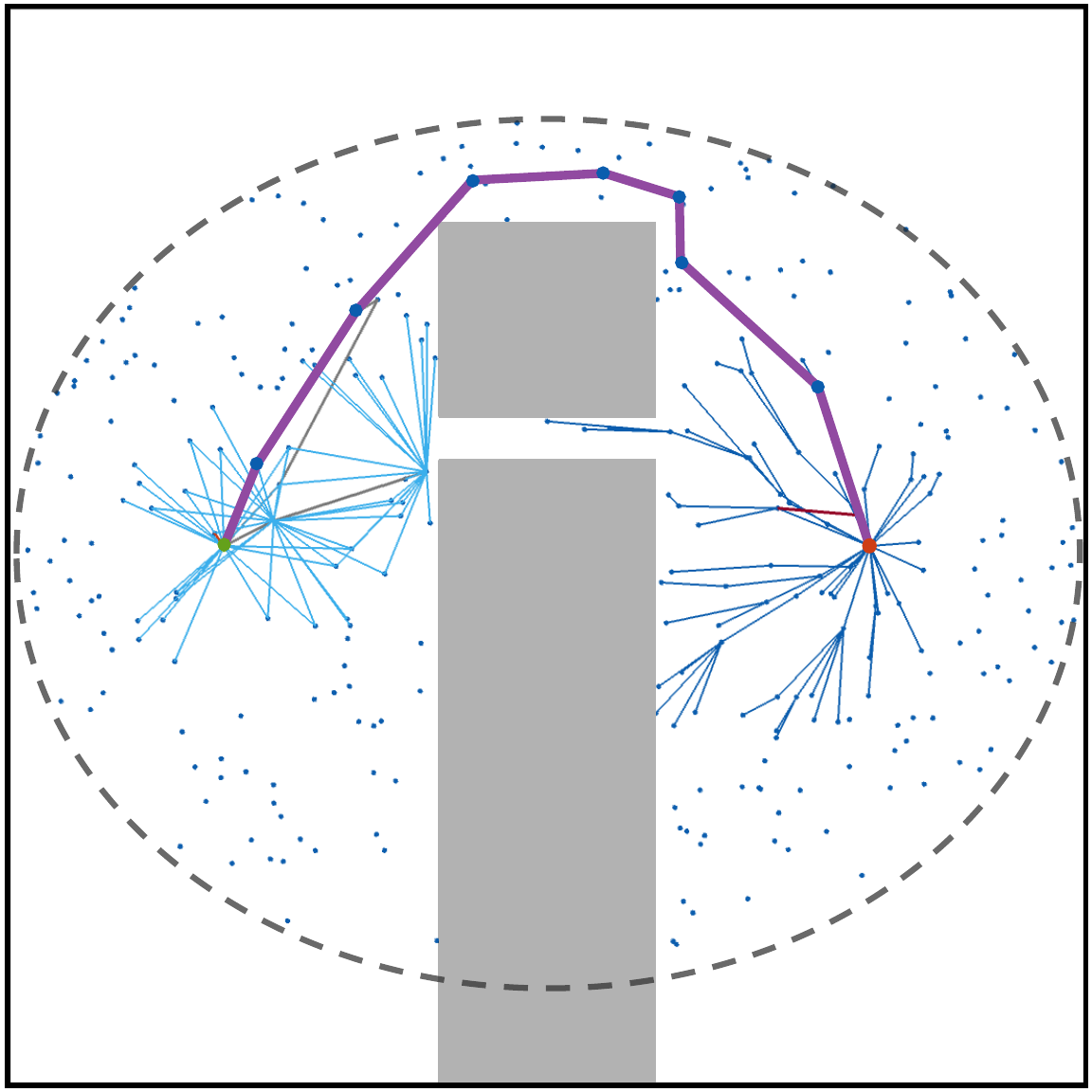}};
    \node[inner sep=0pt] (russell) at (1.4,-5.2)
    {\includegraphics[width=0.24\textwidth]{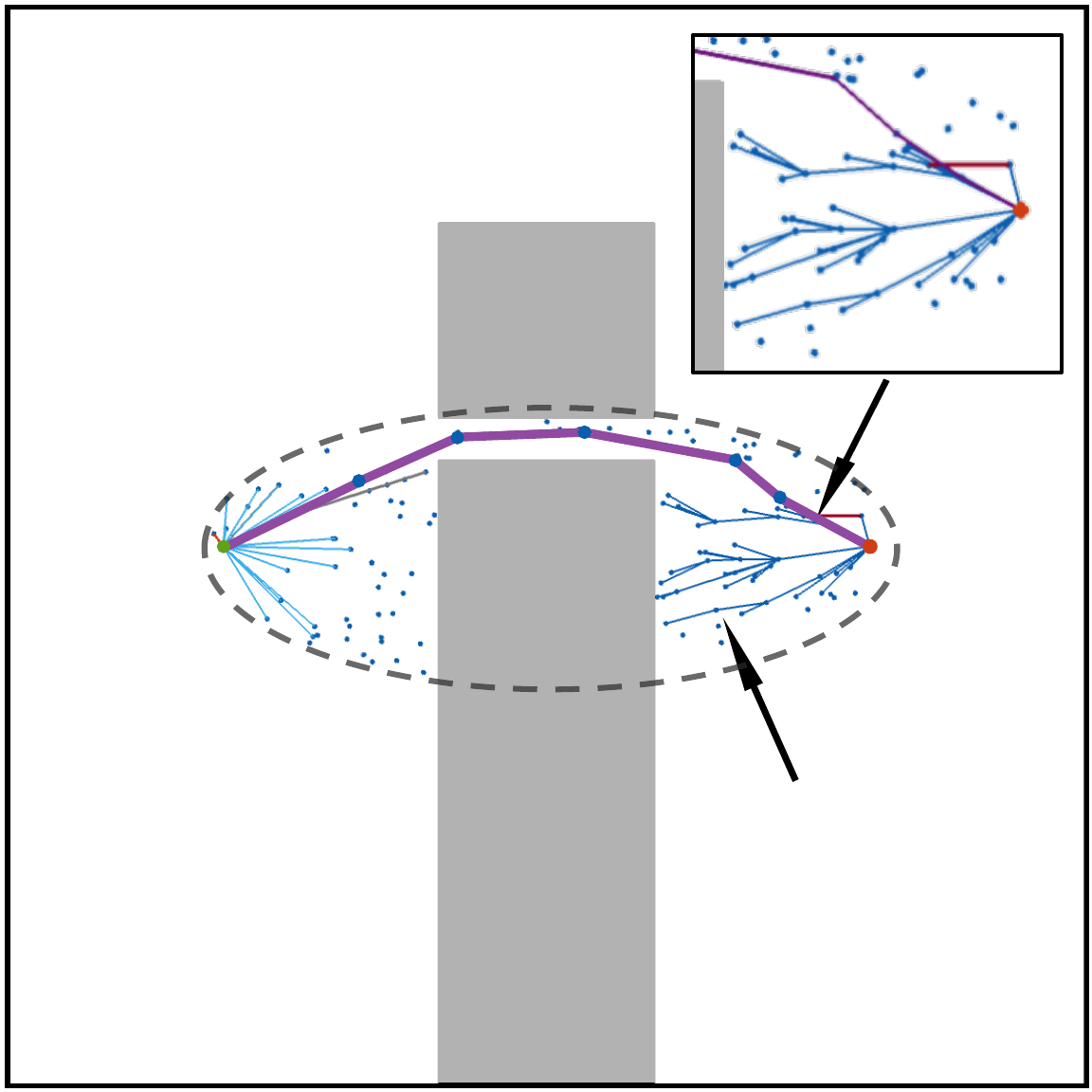}};

    \node at (-1.75,-1.95) {\color{teal}\scriptsize batch size: 100};
    \node at (2.65,-1.95) {\color{teal}\scriptsize batch size: 100};
    \node at (-1.75,-7.12) {\color{purple}\scriptsize batch size: 199};
    \node at (2.65,-7.12) {\color{purple}\scriptsize batch size: 56};

    \node at (2.65,-1.1) {\color{purple}\scriptsize slow edge check};
    \node at (2.65,-6.27) {\color{teal}\scriptsize fast edge check};

    \node at (-3,-2.48) {\small(a) EIT*-initial phase (sparse)};
    \node at (1.4,-2.48) {\small(b) EIT*-optimal phase (dense)};
    \node at (-3,-7.65) {\small(c) FIT*-initial phase (dense)};
    \node at (1.4,-7.65) {\small(d) FIT*-optimal phase (sparse)};
    
    \end{tikzpicture}
    \caption{Four snapshots of how EIT* and FIT* place their sampling strategy. FIT* employs an adaptive batch size (1 to 199) strategy. EIT* maintains a constant batch size (100) across all sample batches. The approach contains a combination of sparse and dense sampling, where more samples are utilized to expedite tree construction (c) compared to (a). Subsequently, FIT* reduces the samples per batch, resulting in less time-consuming edge checking and an increased frequency of batch updating (d) compared to (b).}
    \label{fig: BatchSize}
    \vspace{-1.7em} 
\end{figure}
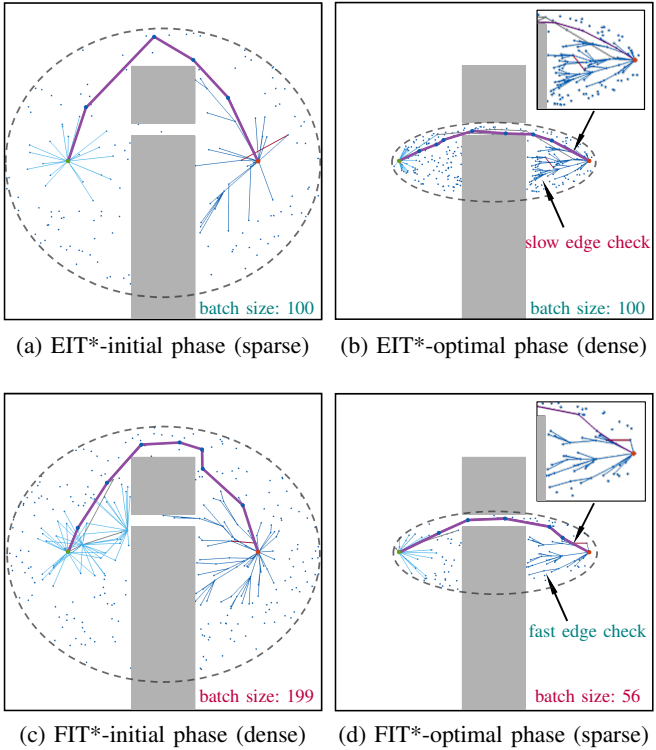

In sampling-based motion planning, two distinct strategies are introduced for sampling vertices in the \textit{$\mathcal{C}$-space}: \textit{Dense sampling} and \textit{Sparse sampling} (Fig.~\ref{fig: BatchSize}). Dense sampling entails generating more samples per batch, thereby increasing the likelihood of including critical vertices within narrow corridors (key sample areas). However, dense sampling demands more time for constructing the RGG, elongating the duration required for edge checking. On the contrary, sparse sampling involves generating fewer samples per batch, thereby expediting the construction of RGG and reducing the time needed for edge checking. Nonetheless, this approach entails a trade-off, as it diminishes the likelihood of sampling vertices within key sample areas. Despite this trade-off, sparse sampling accelerates the edge-checking process, facilitating a quicker transition to subsequent sampling iterations.


    
    

\begin{figure}[t!]
    \centering
    \begin{tikzpicture}
    
    \node[inner sep=0pt] (russell) at (0.0,0.0)
    {\includegraphics[width=0.44\textwidth]{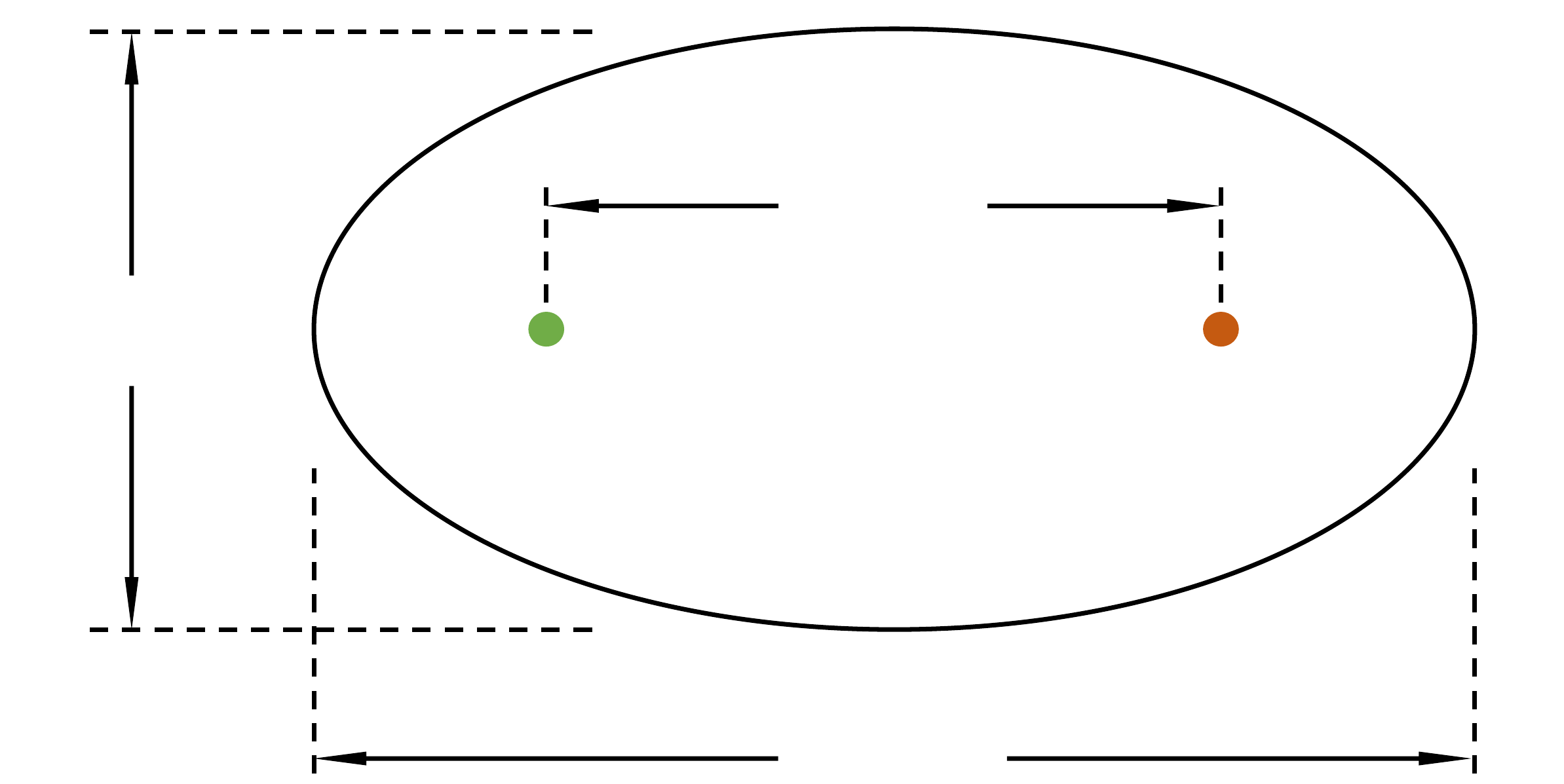}};
    \node at (0.55,0.95) {$l_\text{min}$};
    \node at (0.55,-1.85) {$l_\text{curr}$};
    \node at (-1.2,0.0) {\color{teal} Start};
    \node at (2.2,0.0) {\color{purple} Goal};
    \node at (-3.4,0.28) {$\sqrt{l_\text{curr}^2 - l_\text{min}^2}$};
    
    \node at (0.55,-0.88) {$\color{violet}\textit{informed set}$};
        
    \end{tikzpicture}
    \caption{The 2D representation of hyperellipsoid, its shape is determined by the start, goal states, and two key costs: the theoretical minimum cost $l_\text{min}$ and the current best solution cost $l_\text{curr}$. The eccentricity of the hyperellipsoid is given by the ratio $l_\text{min}/l_\text{curr}$ \cite{gammell2014informed}.}
    \label{fig: elipse}
    \vspace{-1.7em} 
\end{figure}
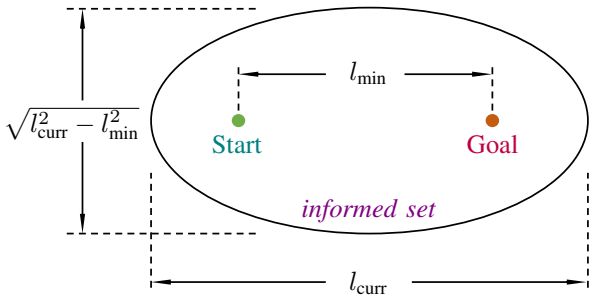

%
%
Effective motion planning involves striking a balance between \textit{exploration} and \textit{exploitation}. Overemphasizing exploration wastes time mapping the environment without progressing toward the goal, while excessive exploitation may overlook superior solutions~\cite{karaman2011sampling}. Exploring/Exploiting Trees (EET)~\cite{rickert2008balancing} aim to strike this balance by integrating gradient information.
RRT-Connect extends RRT to efficiently find a path between two points in $\mathcal{C}$-space by utilizing two RRTs to connect start and goal states.
RRT*~\cite{karaman2011sampling} enhances efficiency and optimality by incrementally constructing a tree from an initial state. Informed-RRT\cite{gammell2014informed} and Smart-RRT~\cite{nasir2013rrt}, extensions of RRT, use heuristics or informed sampling (Fig.~\ref{fig: elipse}) to guide state space exploration, biasing sampling towards regions likely to contain the optimal path. Lazy-PRM~\cite{bohlin2000path} improves roadmap construction by deferring collision checking until necessary.

%
BIT* is a sampling-based planner designed to achieve almost-surely asymptotic optimality. It utilizes a \textit{batch processing approach}, takes numerous state batches and approximates as a progressively denser edge-implicit RGG. It optimizes the tree structure iteratively to minimize computational overhead. 
EIT* and AIT* consist of forward edge build and reverse search to employ distinct heuristic approaches for pathfinding. AIT* excels in precision by updating its heuristic with high accuracy to match the ongoing approximation. EIT* is built on BIT* and uses effort as an additional heuristic function to estimate the collision check of the edges. In contrast, EIT* focuses on optimizing the pathfinding process by utilizing the count (e.g.~\textit{number of collision checks}) as an effort estimate heuristic, leading to efficient path discovery. Through iterative optimization, they approximate a near-optimal path, effectively balancing exploration and exploitation for efficient robot navigation. 
\subsection{Ellipsoid in high-dimensional space}
An ellipsoid in an \( n \)-dimensional Euclidean space \( \mathbb{R}^n \) is defined as the locus of all points \( \mathbf{x} \) that satisfy the equation:
\begin{equation}
    \left( \frac{x_1}{v_1} \right)^2 + \left( \frac{x_2}{v_2} \right)^2 + \cdots + \left( \frac{x_n}{v_n} \right)^2 = 1,
\end{equation}
Where \( \mathbf{x} = (x_1, x_2, \ldots, x_n) \) represents the coordinates of a point on the ellipsoid, and \( \mathbf{v} = (v_1, v_2, \ldots, v_n) \) is a given vector in \( \mathbb{R}^n \). The components of the vector \( \mathbf{v} \) correspond to the lengths of the semi-axes of the ellipsoid along each coordinate axis in \( \mathbb{R}^n \)\cite{han2018eca}.

Open Motion Planning Library (OMPL)~\cite{sucan2012open}, is a widely used open-source database designed to address motion planning challenges in robotics and related fields. It provides a comprehensive framework and a suite of tools to assist researchers and developers in developing efficient motion planning algorithms. We integrated the proposed algorithm, Flexible Informed Trees (FIT*), into the OMPL framework and Planner-Arena benchmark database~\cite{moll2015benchmarking}, along with Planner Developer Tools (PDT)~\cite{gammell2022planner}.
\section{Problem Formulation}
We define the problem of optimal planning in a manner associated with the definition provided in~\cite{karaman2011sampling}.

\textit{Problem Definition 1 (Optimal Planning):} Consider a planning problem with the state space $X \subseteq \mathbb{R}^n$. Let $X_{\text{obs}} \subset X$ represent states in collision with obstacles, and $X_{\text{free}} = cl(X \setminus X_{\text{obs}})$ denote the resulting permissible states, where $cl(\cdot)$ represents the \textit{closure} of a set. The initial state is denoted by $\mathbf{x}_{\text{start}} \in X_{\text{free}}$, and the set of desired final states is $X_{\text{goal}} \subset X_{\text{free}}$. A sequence of states $\sigma: [0, 1] \mapsto X$ forms a continuous map (i.e., a collision-free, feasible path), and $\Sigma$ represents the set of all nontrivial paths.

The optimal solution, represented as $\sigma^*$, corresponds to the path that minimizes a selected cost function $s: \Sigma \mapsto \mathbb{R}_{\geq 0}$. This path connects the initial state $\mathbf{x}_{\text{start}}$ to any goal state $\mathbf{x}_{\text{goal}} \in X_{\text{goal}}$ through the free space:
\begin{equation}
\begin{split}
    \sigma^* &= \arg \min_{\sigma \in \Sigma} \left\{ s(\sigma) \middle| \sigma(0) = \mathbf{x}_{\text{start}}, \sigma(1) \in \mathbf{x}_{\text{goal}}, \right. \\
    &\qquad\qquad \left. \forall t \in [0, 1], \sigma(t) \in X_{\text{free}} \right\},
\end{split}
\end{equation}
Where $\mathbb{R}_{\geq 0}$ denotes non-negative real numbers. The cost of the optimal path is $s^*$.

Considering a discrete set of states, $X_{\text{samples}} \subset X$, as a graph where edges are determined algorithmically by a transition function, we can describe its properties using a probabilistic model implicit dense RGGs when these states are randomly sampled, i.e., $X_{\text{samples}} = \{ \mathbf{x} \sim \mathcal{U}(X) \}$, as discussed in~\cite{penrose2003random}.

The characteristics of the anytime almost-surely sampling-based planner with the definition are provided in~\cite{gammell2018informed}.

\textit{Problem Definition 2 (Almost-sure asymptotic optimality):} A planner is considered almost-surely asymptotically optimal as the number of samples tends to \textit{infinity} which covers the entire general \textit{$\mathcal{C}$-space}. In this scenario, if an optimum solution exists (as definition 1 optimal planning), the probability of the planner asymptotically converging to the optimal solution equals one when sample size $q$ goes infinite.

\begin{equation}
     P \left( \limsup_{q \to \infty} c(\sigma_q) = c(\sigma^*) \right) = 1.
\end{equation}
where $q$ represents the number of samples a planner has sampled, $\sigma_q$ signifies the path derived by the planner from those batch of samples, $\sigma^*$ stands for the optimal solution to the planning problem, and $c(\cdot)$ denotes the path's cost at the informed batch. After discovering an initial solution, the planner utilizes the remaining computational time to optimize the path quality of the existing solution.

\section{Flexible Informed Trees (FIT*)}



FIT* builds on EIT* and follows the asymmetric bidirectional search process with adaptive batch-size tunning (Alg.~\ref{biSearch}).
Moreover, FIT* dynamically adjusts batch sizes based on the state space's geometry, fine-tuning the sampling density according to the dimension of \textit{$\mathcal{C}$-space} and hypervolume of the $n$-dimensional hyperellipsoid. This difference in sampling strategy leads to distinct efficiencies and computational performance within planning algorithms.

\begin{algorithm}[t!]
\caption{Asymmetric bidirectional search}
\label{biSearch}
\small
\DontPrintSemicolon
\SetKwIF{If}{ElseIf}{Else}{if}{}{else if}{else}{end if}%
\Repeat{planner termination condition}{ 
\emph{improve RGG approximation (re-sampling)}\\
\emph{\textcolor{purple}{update current batch size (adaptive adjust)}}\\
\emph{calculate heuristics (reverse search)}\\

\While{RGG approximation is beneficial}{
    \emph{discover the feasible path (forward search)}\\
    \If{found invalid edge or unprocessed state}{
        \emph{update heuristics (reverse search)}}}
}
\end{algorithm}

\subsection{Notation}~\label{subsec: notation}
The state space of the planning problem is denoted by $X \subseteq \mathbb{R}^n$, where $n \in \mathbb{N}$. The start point is represented by $\mathbf{x}_{\text{start}} \in X$, and the goals are denoted by $X_{\text{goal}} \subset X$. The sampled states are denoted by $X_{\text{sampled}}$. The forward and reverse search trees are represented by $\mathcal{F} = (V_\mathcal{F}, E_\mathcal{F})$ and $\mathcal{R} = (V_\mathcal{R}, E_\mathcal{R})$, respectively. The vertices in these trees, denoted by $V_\mathcal{F}$ and $V_\mathcal{R}$, are associated with valid states. The edges in the forward tree, $E_\mathcal{F} \subset V_\mathcal{F} \times V_\mathcal{F}$, represent valid connections between states, while the edges in the reverse tree, $E_\mathcal{R} \subset V_\mathcal{R} \times V_\mathcal{R}$, may traverse invalid regions of the problem domain. An edge comprises a source state, $\mathbf{x}_s$, and a target state, $\mathbf{x}_t$, denoted as $(\mathbf{x}_s, \mathbf{x}_t)$. $\mathcal{Q_F}$ and $\mathcal{Q_R}$ designate the edge-queue for the forward search and reverse search, respectively. The true connection cost between two states in \textit{$\mathcal{C}$-space} is represented by the function $c: X \times X \rightarrow [0, \infty)$.


For sets $A, B,$ and $C$ with $B, C$ being subsets of $A$, the notation $B \stackrel{+}{\leftarrow} C$ denotes $B \leftarrow B \cup C$, and $B \stackrel{-}{\leftarrow} C$ denotes $B \leftarrow B \setminus C$.

\textit{FIT*-specific Notation:}
Incorporating attenuation introduces a decay factor $\Psi_\text{decay}$, with specified minimal $m_\text{min}:= 1$ and maximal $m_\text{max}$ sample numbers per batch. The initial batch sizes, denoted as $\mathcal{M}_\text{initial}$, and the current count of states sampled per batch represented as $\mathcal{M}(\Psi_\text{current})$. The Lebesgue measure within an $n$-dimensional hyperellipsoid is denoted by $\zeta_n$. Non-negative scalar $\xi_n$ represents the raw ratio of the initial hypervolume of the $n$-dimensional hyperellipsoid $\mathcal{V}_\text{initial}$ to the current hypervolume $\mathcal{V}_\text{current}$. A nature logarithmic tuning 
parameter $\Lambda_\text{tuning}$ regulates the rate at which the ratio decreases. $\mathcal{O}_\text{smooth}$ represents a smoothed value after the attenuation of the initial and optimal state.

\begin{algorithm}[t!]
\caption{Flexible Informed Trees (FIT*)}
\label{FIT_algori}
\DontPrintSemicolon
\SetKwIF{If}{ElseIf}{Else}{if}{}{else if}{else}{end if}%
\SetKwFunction{expand}{expand}
\SetKwFunction{parent}{parent}
\SetKwFunction{updateDecayFactor}{updateDecayFactor}
\SetKwFunction{sigmoidFunction}{sigmoidFunction}
\SetKwFunction{updateRawRatio}{updateRawRatio}
\SetKwFunction{getBestForwardEdge}{getBestForwardEdge}
\SetKwFunction{collisionFree}{collisionFree}
\SetKwFunction{adaptiveBactchSize}{adaptiveBactchSize}
\SetKwFunction{prune}{prune}
\SetKwFunction{sample}{sample}
\DontPrintSemicolon
\scriptsize

\emph{$c_\text{current} \leftarrow \infty; \Psi_\text{decay} \leftarrow \infty; \Lambda_\text{tuning} \leftarrow \text{initialized}$}\\ 
\emph{$\mathcal{M}_\text{initial} \leftarrow \mathcal{M}(\Psi_\text{current}); \mathcal{V}_\text{current}\leftarrow \mathcal{V}_\text{initial};\xi_\text{initial}:= 1 $}
\Comment{initial batch size}\\
\emph{$X_\text{sampled} \leftarrow X_\text{goal} \cup \{ \mathbf{x}_\text{start} \} $}\\ 
\emph{$V_\mathcal{F}, E_\mathcal{F}, \mathcal{Q_F} \leftarrow \text{\expand}(\mathbf{x}_\text{start})$}\\ 
\emph{$V_\mathcal{R}, V_{\mathcal{R},\text{closed}}, E_\mathcal{R}, \mathcal{Q_R} \leftarrow \text{\expand}(X_\text{goal})$}\\ 
\emph{$\text{Initialize and update the Inflation Factor and Sparse Resolution}$}\\

\Repeat{planner termination condition}{
    \emph{$\xi_n \leftarrow \text{\updateRawRatio}(\mathcal{V}_\text{current},\mathcal{V}_\text{initial})$}\\ 
    \emph{$\mathcal{O}_\text{smooth} \leftarrow \text{\sigmoidFunction}(\xi_n)$}
    \Comment{update smoothed raw ratio}\\
    \emph{$\Psi_\text{decay} \leftarrow \text{\updateDecayFactor}(\mathcal{O}_\text{smooth}, \Lambda_\text{tuning})$}\\
    \uIf{\text{Forward queue has better edge priority than Reverse queue} \\$\mathbf{or} \text{ target of optimal edge in forward queue}$}{
        \emph{$(\mathbf{x}_s, \mathbf{x}_t) \leftarrow \text{{argmin}}_{(\mathbf{x}_s, \mathbf{x}_t) \in \mathcal{Q_R}} \{\text{{key}}^\text{{FIT*}}_\mathcal{R}(\mathbf{x}_s, \mathbf{x}_t)\}$}\\
        \Comment{heuristics of solution cost and effort with batch-size tuning}\\
        \emph{$\mathcal{Q_R} \stackrel{-}{\leftarrow} (\mathbf{x}_s, \mathbf{x}_t) $}\\
        \emph{$V_{\mathcal{R},\text{closed}} \stackrel{+}{\leftarrow} \mathbf{x}_s$}\\
        \uIf{\text{{No sparse collisions detected between }} $(\mathbf{x}_s, \mathbf{x}_t)$}{
            \emph{\text{{Update cost and reverse tree structures}}}
            \emph{\color{purple}${\mathcal{M}(\Psi_\text{current})\leftarrow \adaptiveBactchSize(\Psi_\text{decay}, m_\text{min}, m_\text{max})}$}\\ 
        }\Else{
            \emph{\text{{Execute sparse collision case processing}}}

        }
    }\uElseIf{\text{{Better edge in forward queue compared to current solution}}}{
        \emph{$(\mathbf{x}_s, \mathbf{x}_t) \leftarrow \text{{\getBestForwardEdge}}(\mathcal{Q_F})$}\\
        \emph{$\mathcal{Q_F} \stackrel{-}{\leftarrow} (\mathbf{x}_s, \mathbf{x}_t)$}\\
        \uIf{\text{{Edge is in existing forward tree}}}{
             \emph{$\xi_n \leftarrow \text{\updateRawRatio}(\mathcal{V}_\text{current},\mathcal{V}_\text{initial})$}\\ 
            \emph{$\mathcal{O}_\text{smooth} \leftarrow \text{\sigmoidFunction}(\xi_n)$}\\ 
            \emph{\text{{Update forward tree structures}}}
        }\ElseIf{\text{{Edge improves existing path}}}{
            \uIf{\text{{No dense collisions between }} $(\mathbf{x}_s, \mathbf{x}_t)$}{
                \emph{\text{{Update cost and forward tree structures}}}
                \emph{$\Psi_\text{decay} \leftarrow \text{\updateDecayFactor}(\mathcal{O}_\text{smooth}, \Lambda_\text{tuning})$}\\
            }\Else{
                \emph{\text{{Execute dense collision case processing}}}\\
            }
        }
    }\Else{
        \emph{\text{{Prune sampled states}}}
        \emph{\color{purple}${\mathcal{M}(\Psi_\text{current})\leftarrow \adaptiveBactchSize(\Psi_\text{decay}, m_\text{min}, m_\text{max})}$}\\ 
    }
}
\end{algorithm}


\subsection{Approximation}\label{subsec: approx.}
FIT* employs informed sampling strategies \cite{gammell2018informed} to concentrate its RGG approximation on the pertinent region of the state space and dynamically adjusts the connection radius as more states are sampled. The radius (\( r \)) is updated according to the approach proposed in \cite{karaman2011sampling}, utilizing the measure of the informed set as introduced in \cite{gammell2018informed}.
\begin{equation}
\label{eqn:radius r}
    r(q) = \eta \left(2 \left(1 + \frac{1}{n}\right){\left(\frac{\lambda(X_{\hat{f}})}{\zeta_n}\right) \left( \frac{\log(q)}{q}\right)}\right)^{\frac{1}{n}},
\end{equation}
Here, $q$ denotes the number of sampled states in the informed set, $\eta >$ 1 is a tuning parameter, $\lambda(\cdot)$ denotes the Lebesgue measure, and $n$ represents the dimension of the state space.
\begin{algorithm}[t!]
\caption{\small\text{adaptiveBactchSize} $(\Psi_\text{decay}, m_\text{min}, m_\text{max})$}
\label{adaptiveBatch}
\DontPrintSemicolon
\scriptsize
\SetKwIF{If}{ElseIf}{Else}{if}{}{else if}{else}{end if}%
\SetKwFunction{updateDecayFactor}{updateDecayFactor}
\SetKwFunction{sigmoidFunction}{sigmoidFunction}
\SetKwFunction{updateRawRatio}{updateRawRatio}
\SetKwFunction{collisionFree}{collisionFree}
\SetKwFunction{ellipseVolumCal}{ellipseVolumCal}
\SetKwFunction{updateSolutionCost}{updateSolutionCost}

\emph{$m_\text{min}:= 1, m_\text{max} := 2m_\text{current} - m_\text{min}$}\\
\If{$c_\text{current} \neq c_\text{last}\;\mathbf{or}\;c_\text{last} := \infty$}
{
    \If{$c_\text{current} := \infty$}{\Return{null}\Comment{ensure safety when initial solution not found yet}}
    \emph{$c_\text{last} \leftarrow$ \updateSolutionCost($c_\text{current}$)}\\
    \If{pragma once}{\emph{$\mathcal{V}_\text{initial} \stackrel{+}{\leftarrow} \ellipseVolumCal (c_\text{last})$}\Comment{$\mathcal{V}_\text{initial}$ only calculated once}\\}

    \emph{$\mathcal{V}_\text{current} \stackrel{+}{\leftarrow} \ellipseVolumCal(c_\text{current})$} 
    \Comment{$\mathcal{V}_\text{current}$ update every-time }\\
    \emph{$\xi_n \leftarrow \updateRawRatio(\mathcal{V}_\text{current},\mathcal{V}_\text{initial})$}\\
    \emph{$\mathcal{O}_\text{smooth} \leftarrow \sigmoidFunction(\xi_n)$}
    \Comment{equation (\ref{fuc:smooth})}\\
    \emph{$\Psi_\text{decay} \leftarrow \updateDecayFactor(\mathcal{O}_\text{smooth}, \Lambda_\text{tuning})$}
    \Comment{equation (\ref{fuc:decay})}\\
    \emph{$\mathcal{M}(\Psi_\text{current}):= m_\text{min}+\Psi_\text{decay}(m_\text{max}-m_\text{min})$}
    \Comment{adapted batch size}\\
    \BlankLine
    \Return{$\mathcal{M}(\Psi_\text{current})$}
}

\end{algorithm}

\begin{figure}[t!]
    \centering
    \begin{tikzpicture}
    
    \node[inner sep=0pt] (russell) at (0.0,0.0)
    {\includegraphics[width=0.35\textwidth]{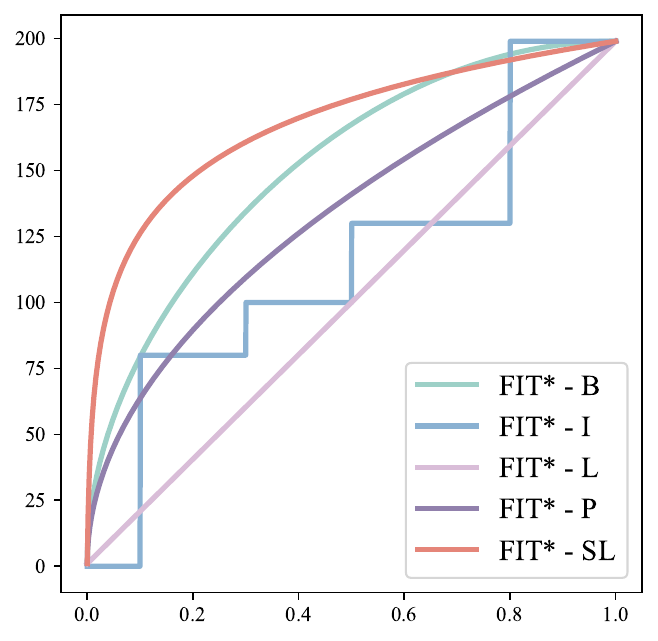}};

    \node [rotate=90] at (-3.4,0.0) {\small\textit{Samples per batch }$\mathcal{M}(\Psi)$};
    
    \node at (0.0,-3.28) {\small\textit{Decay factor }$\Psi_\text{decay}$};
        
    \end{tikzpicture}
    \caption{This graph shows the decay-based method comparison, where the maximal batch size is 199, and the minimal batch size is 1; more specific illustrations are given in section~\ref{subsec: adaBachSize}.}
    \label{fig: decay_method}
    \vspace{-1.7em} 
\end{figure}
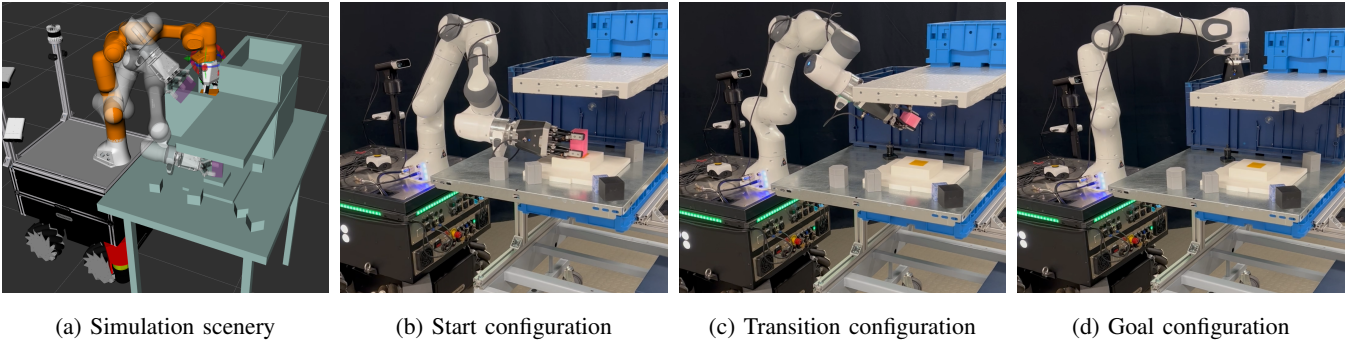
\begin{figure*}[t!]
    \centering
    \begin{tikzpicture}
    \footnotesize
    \node[inner sep=0pt] (russell) at (-8.0,0)
    {\includegraphics[width=0.244\textwidth]{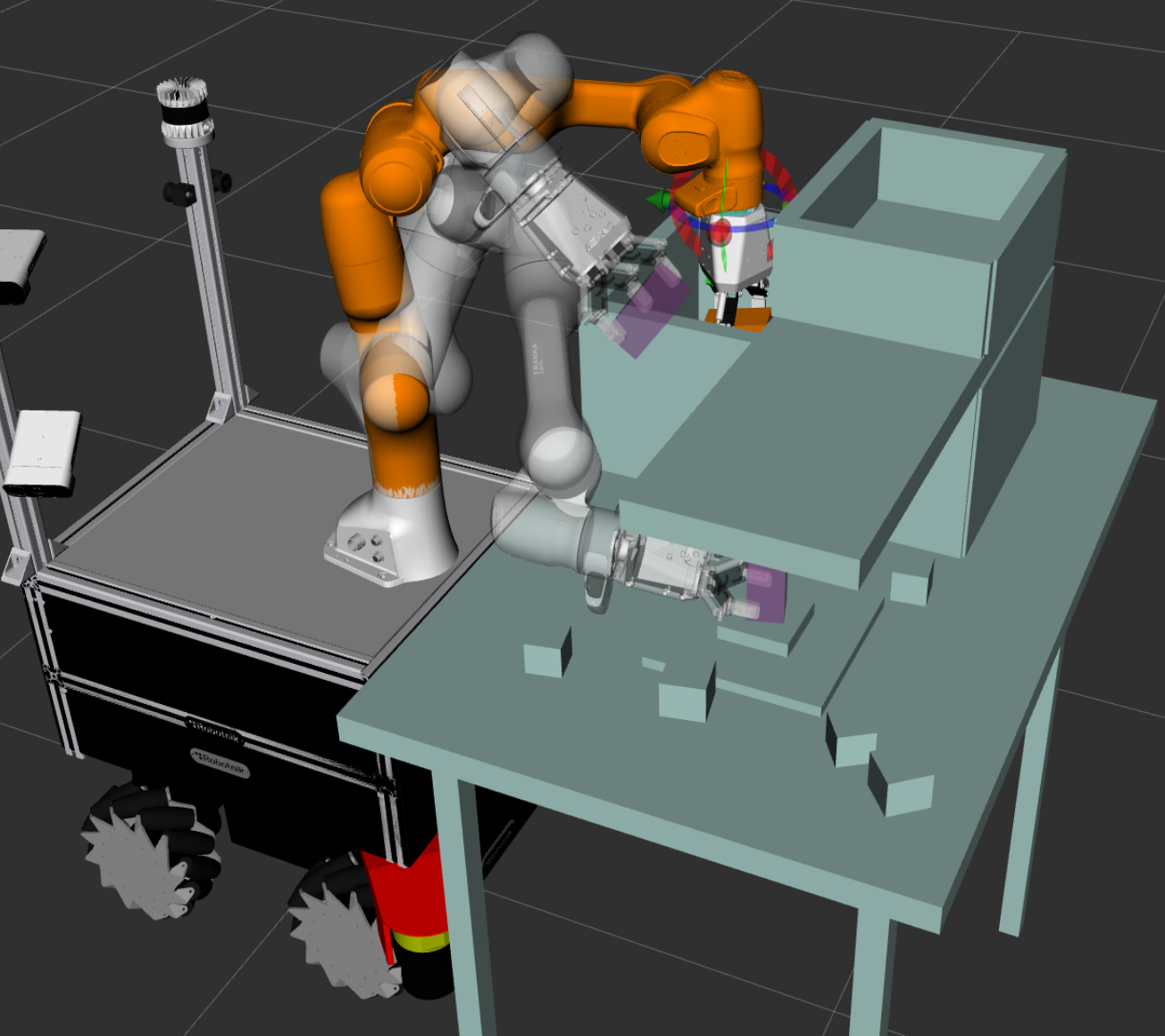}};
    \node[inner sep=0pt] (russell) at (-3.5,0)
    {\includegraphics[width=0.244\textwidth]{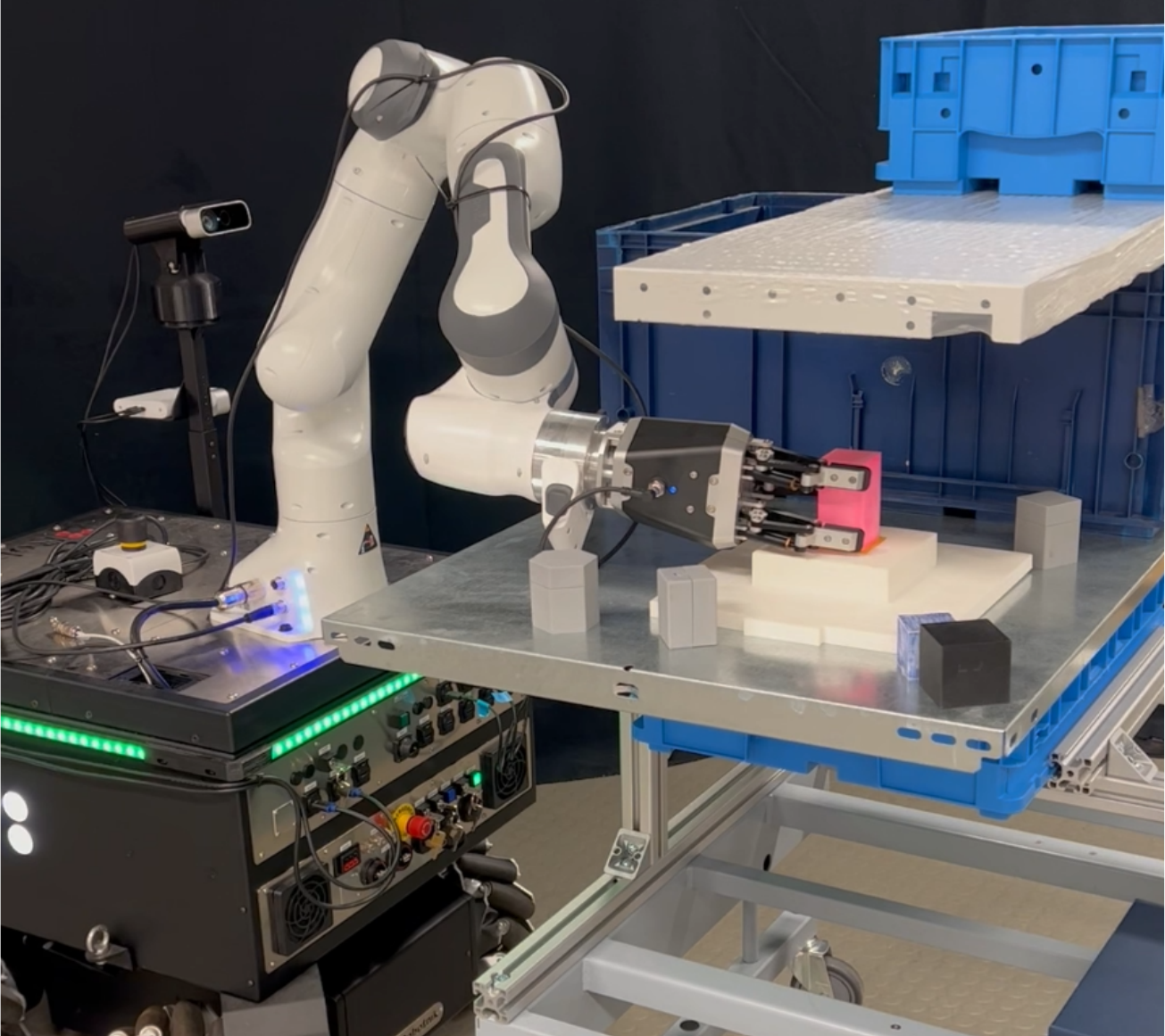}};
    \node[inner sep=0pt] (russell) at (1.0,0)
    {\includegraphics[width=0.244\textwidth]{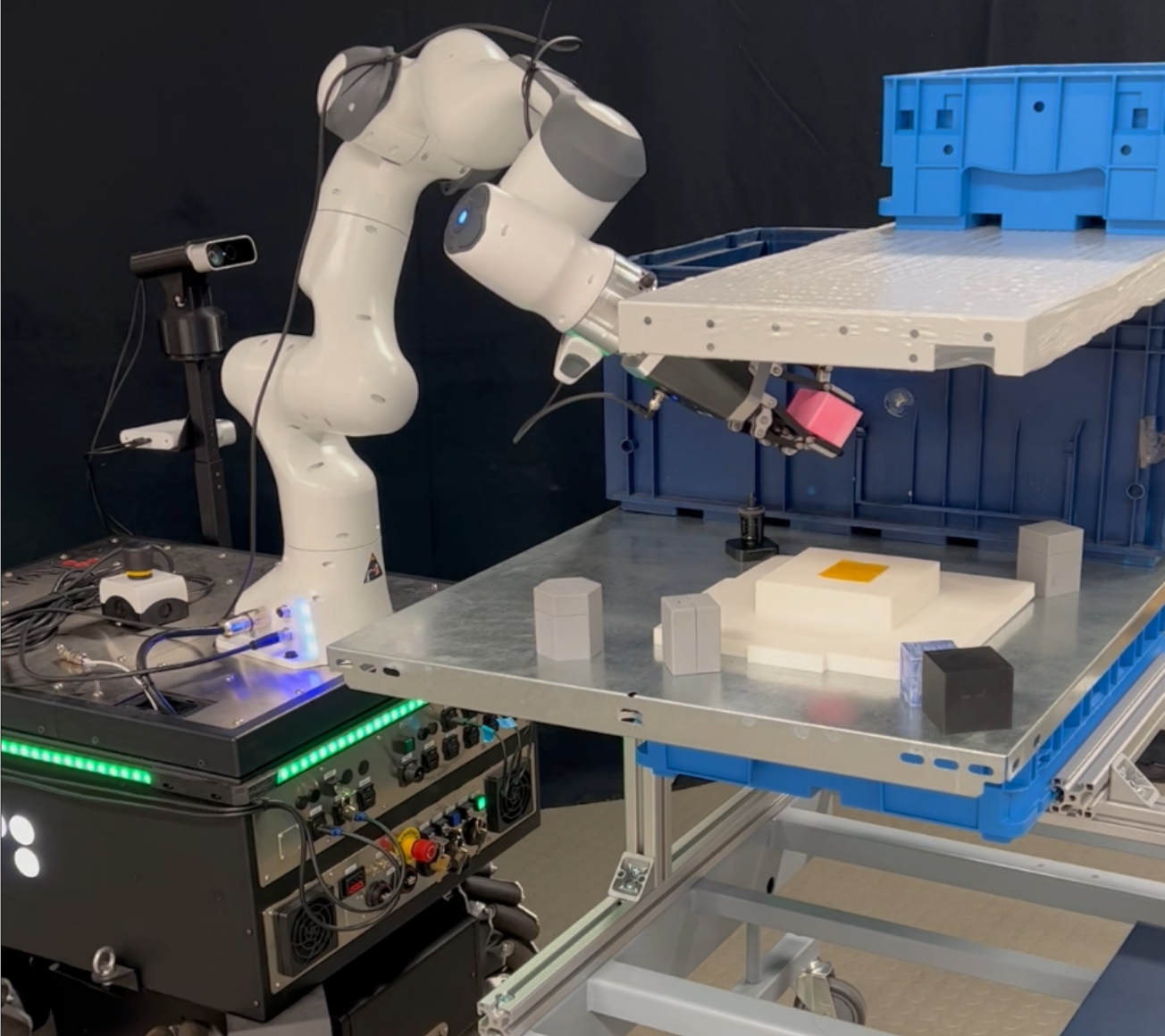}};
    \node[inner sep=0pt] (russell) at (5.5,0)
    {\includegraphics[width=0.244\textwidth]{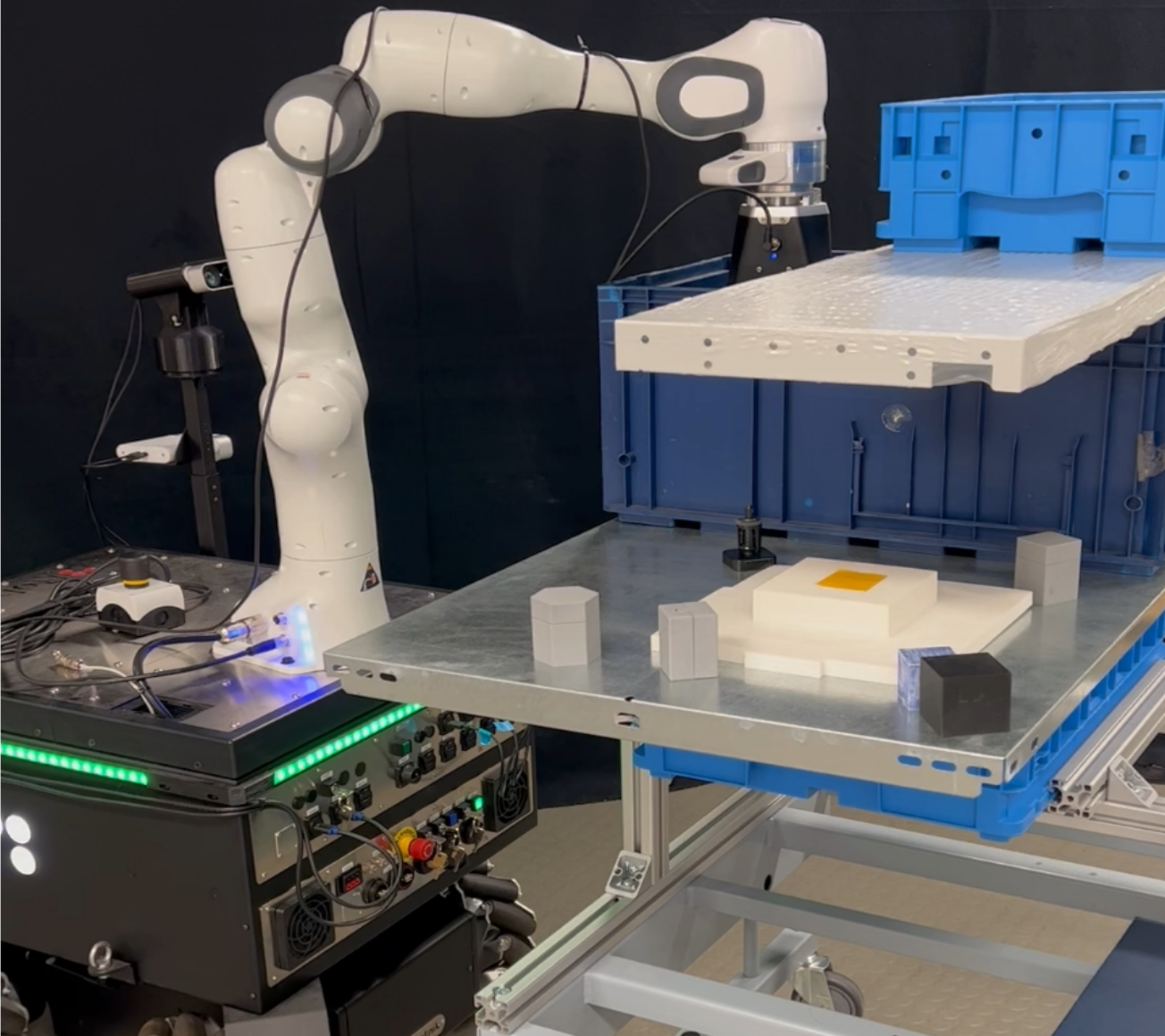}};
    
    \node at (-8.0,-2.4) {\small (a) Simulation scenery};
    \node at (-3.5,-2.4) {\small (b) Start configuration};
    \node at (1.0,-2.4) {\small (c) Transition configuration};
    \node at (5.5,-2.4) {\small (d) Goal configuration};
    
    
    \end{tikzpicture}
    \caption{Illustrates the simulation (a) and the real-world scenarios of DARKO robot for the intralogistics task, (b) shows the start configuration of the arm in position to pick up the red cube from the metal sheet table, (c) shows the transition configuration position of the task, (d) shows the goal configuration of the arm in position to place a cube in the box.}
    \label{fig:simulation}
    \vspace{-1.5em}
\end{figure*}
\subsection{Adaptive Batch-Size}\label{subsec: adaBachSize}
FIT* dynamically adjusts the number of samples in batch size. Specifically, it leverages the sigmoid function in conjunction with logarithmic (\textit{sigmoid-log}, FIT*-SL).
As observed in Table \ref{tab: decay_method}, where $t^\textit{min}_\textit{init}$ represents the minimal initial planning time over 100 runs, and $c^\textit{max}_\textit{init}$ represents the maximal initial cost of the planning problem, respectively. The $c$ and $t$ of unsuccessful attempts are represented as infinity.

\begin{table}[ht]
\caption{Decay-based tuning methods comparison (100 runs)}
\centering
\resizebox{0.485\textwidth}{!}{
\begin{tabular}{p{1.25cm}||c|c|c|c|c|c|c}
 \hline
 & \multicolumn{3}{c}{initial time} & \multicolumn{3}{|c|}{initial cost} &\multirow{2}*{\textbf{success}}\\
    &$t^\textit{min}_\textit{init}$ &$t^\textit{med}_\textit{init}$ &$t^\textit{max}_\textit{init}$ &$c^\textit{min}_\textit{init}$ &$c^\textit{med}_\textit{init}$ &$c^\textit{max}_\textit{init}$ \\
 \hline
    $\textit{EIT*}$    &0.0027 &0.0048 &$\infty$ &0.6526 &1.0442 &$\infty$  &0.97\\
    $\textit{FIT*-B}$  &0.0027  &0.0035  &0.0083 &\textcolor{purple}{0.6333} &0.6747 &1.2486 &\textcolor{purple}{1.00}  \\
    $\textit{FIT*-I}$   &0.0031   &0.0049   &$\infty$ &0.6471 &0.7459 &$\infty$  &0.97  \\
    $\textit{FIT*-L}$   &0.0028   &0.0037   &$\infty$ &0.6441 &0.6956 &$\infty$ &0.99  \\
    $\textit{FIT*-P}$   &0.0028   &0.0034   &0.0077 &0.6406 &0.6753 &1.2502  &\textcolor{purple}{1.00}  \\
    $\textcolor{purple}{\textit{\textbf{FIT*-SL}}}$   
    &\textcolor{purple}{0.0026} &\textcolor{purple}{0.0030} &\textcolor{purple}{0.0062} &\textcolor{purple}{0.6333} &\textcolor{purple}{0.6603} &\textcolor{purple}{1.0608}  &\textcolor{purple}{1.00}\\
 \hline
\end{tabular}} \label{tab: decay_method}
\end{table}

Within the FIT*'s flexible batch size context (Alg.~\ref{FIT_algori} and~\ref{adaptiveBatch}), our comparison focused on exploring various decay methods to comprehend their distinct impacts (Table~\ref{tab: decay_method} and Fig.~\ref{fig: decay_method}). These decay methods encompassed \textit{linear} decay (FIT*-L), \textit{brachistochrone} curve-based decay (FIT*-B), decay functions following an \textit{parabola} pattern (FIT*-P), and decay based on the \textit{iteration count} (FIT*-I). Each method was thoroughly examined 100 runs to assess its effectiveness in shaping the decay factor $\Psi_\text{decay}$. Through experimentation, it is concluded that FIT*-SL stands out as the most efficient approach, demonstrating the quickest \textit{median time} and shortest \textit{median cost} for the initial solution. It strikes a balance between adaptability to denser sampling in the initial pathfinding phase and sparse sampling during the optimization phase of $\mathcal{C}$-space.



\section{Formal Analysis}~\label{sec: formalAnalysis}
In this paper, we refer to Definition 24 from~\cite{karaman2011sampling} to establish the concept of almost-sure asymptotic optimality. 

\subsection{Almost-Sure Asymptotically Optimal Path}
The FIT* algorithm utilizes an RGG approximation similar to EIT*, with the underlying graph likely to contain an asymptotically optimal path. The graph search in FIT* shows asymptotic resolution optimality. Given EIT*'s status as an almost-surely asymptotically optimal algorithm~\cite{gammell2022planner}, it indicates that this RGG approximation includes an asymptotically optimal path. Therefore, the adaptive batch size computation affects the number of edge collision checks and does not seem to impact almost-sure asymptotic optimality.
\begin{figure}[t!]
    \centering
    \begin{tikzpicture}
    \node[inner sep=0pt] (russell) at (-4.0,0.0)
    {\includegraphics[width=0.24\textwidth]{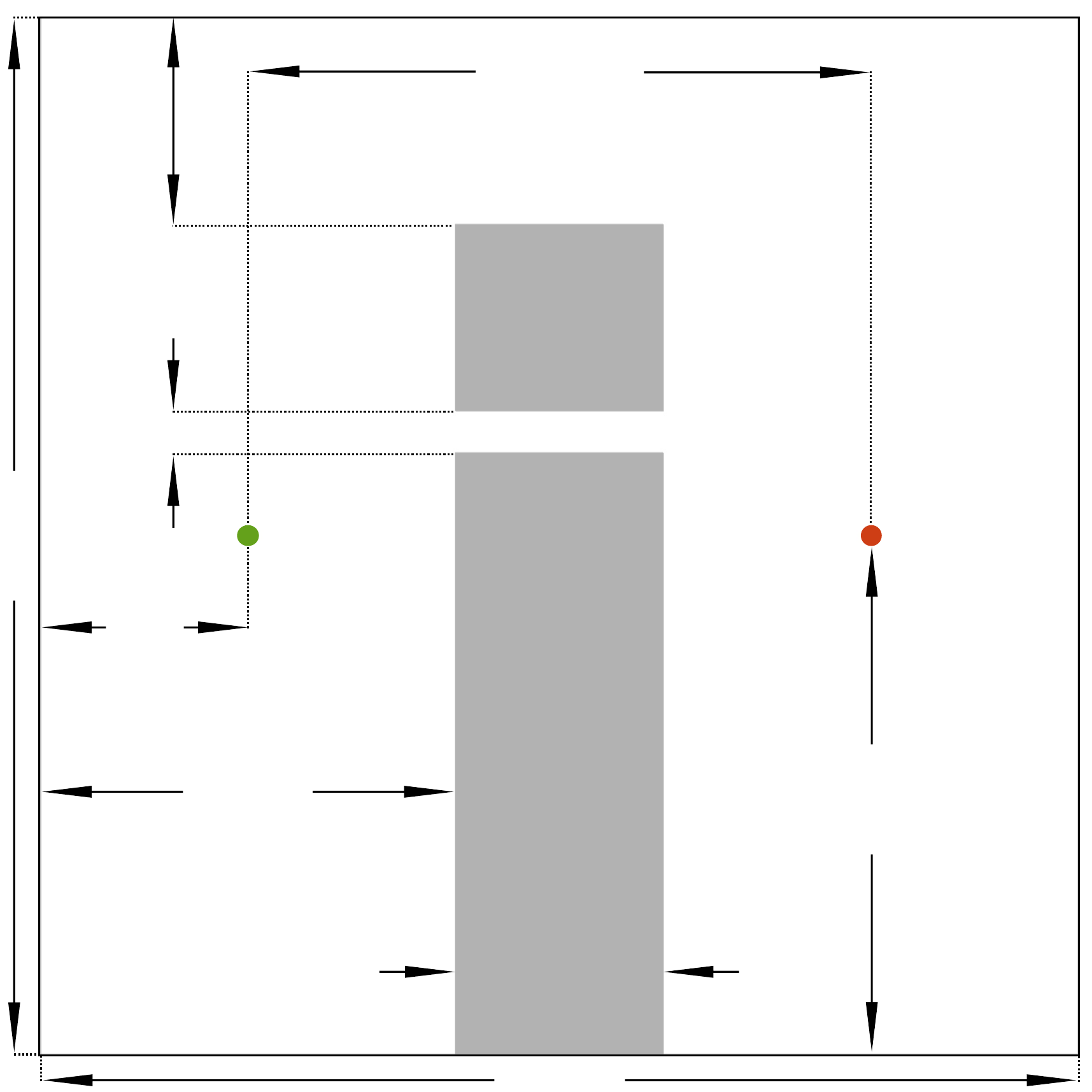}};
    \node[inner sep=0pt] (russell) at (0.25,0.0)
    {\includegraphics[width=0.24\textwidth]{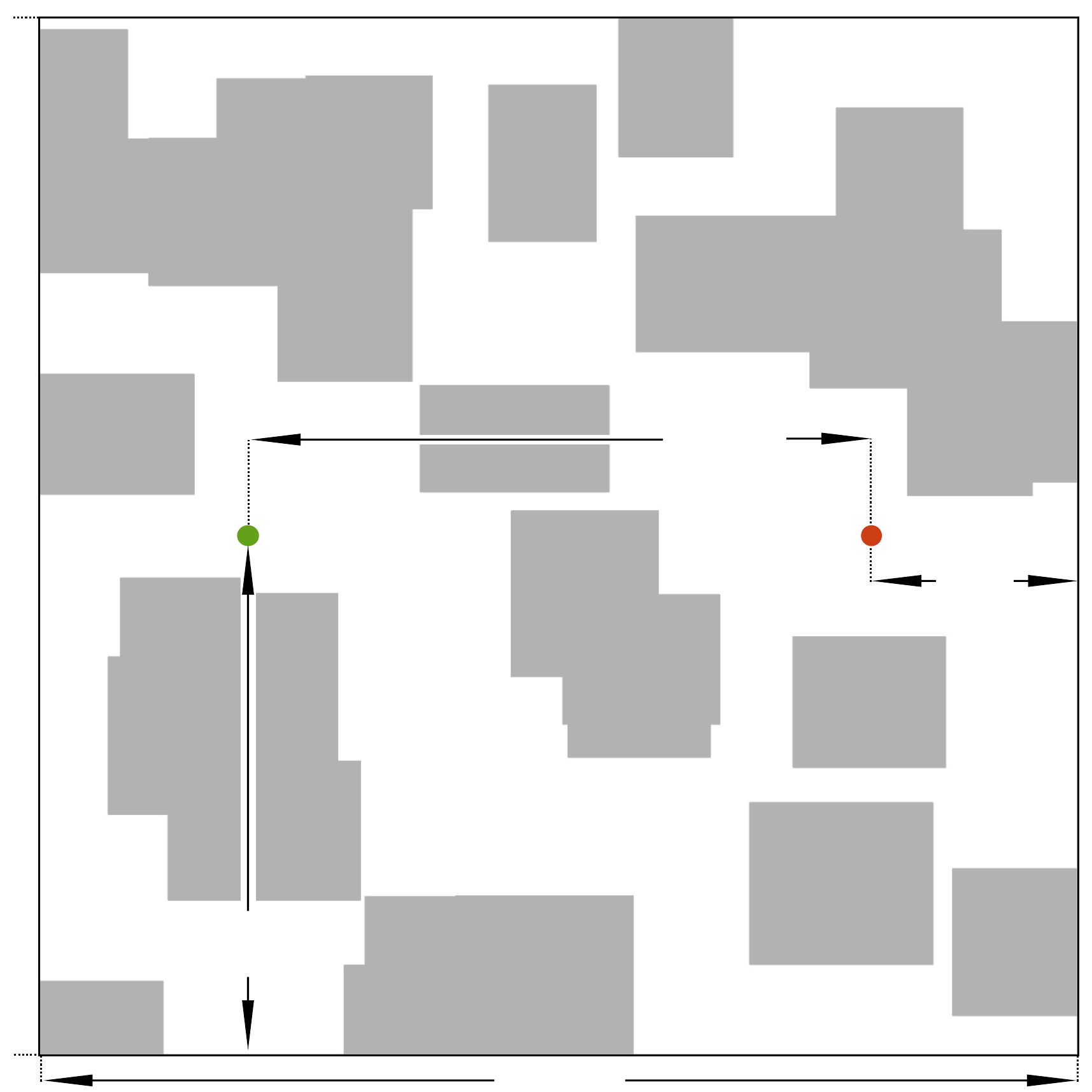}};
    \scriptsize
    
    \node at (-3.97,1.87) {0.6};
    \node at (-5.56,-0.3) {0.2};
    \node [rotate=90] at (-5.6,0.46) {0.04};    
    \node [rotate=90] at (-5.6,1.65) {0.2};
    \node [rotate=90] at (-2.75,-1.0) {0.5};
    \node at (-5.16,-0.97) {0.4};
    \node at (-4.82,-1.65) {0.2};
    \node [rotate=90] at (-6.12,0.05) {1.0};
    \node at (-4.0,-2.13) {1.0};
    \node at (-4.8,0.1) {\color{teal} Start};
    \node at (-3.1,0.1) {\color{purple} Goal};
    
    \node at (0.92,0.43) {0.6};
    \node at (1.92,-0.14) {0.2};
    \node [rotate=90] at (-0.92,-1.54) {0.5};
    \node at (0.25,-2.13) {1.0};
    \node at (-0.5,0.1) {\color{teal} Start};
    \node at (1.14,0.1) {\color{purple} Goal};

    \node at (-4.0,-2.51) {\small(a) Wall Gap};
    \node at (0.25,-2.51) {\small(b) Random Rectangles};
    \end{tikzpicture}
    \caption{The 2D representation of the simulated planning problems in Section~\ref{sec:Expri}. The state space, denoted as $X \subset \mathbb{R}^n$, is constrained within a hypercube with one width for both problem instances. Specifically, we conducted ten distinct instantiations of the random rectangles experiment and the outcomes are showcased in Fig.~\ref{fig: result}.}
    \label{fig: testEnv}
    \vspace{-1.7em} 
\end{figure}
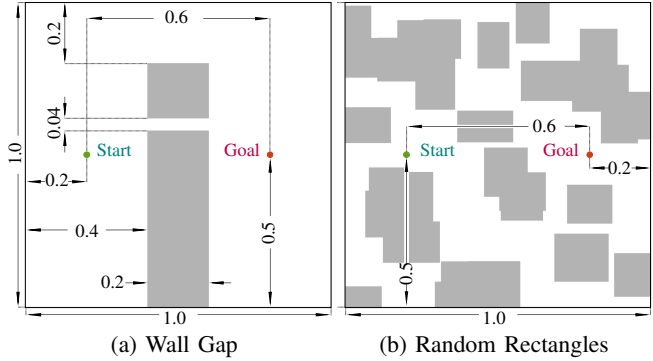
\subsection{Statistical Formulation and Hypotheses}
    The sampling-based planner FIT* dynamically adjusts batch sizes, increasing samples in the initial solution stage for faster convergence to a feasible path. Conversely, reducing samples per batch expedites optimization by minimizing edge checks, accelerating the batch sample update frequency, and reducing computational time.
    The \textit{adaptiveBatchSize} function $\mathcal{M}(\Psi)$ is the major influence, considering the integration of the decay factor. It reflects how the expansion and contraction of the hyperellipsoids in different dimensions affects the appropriate batch size, aligning with the exploration requirements.
    \begin{equation}
        \mathcal{M}(\Psi) := m_\text{min} + (m_\text{max}-m_\text{min}) \times \Psi_\text{decay},
    \end{equation}

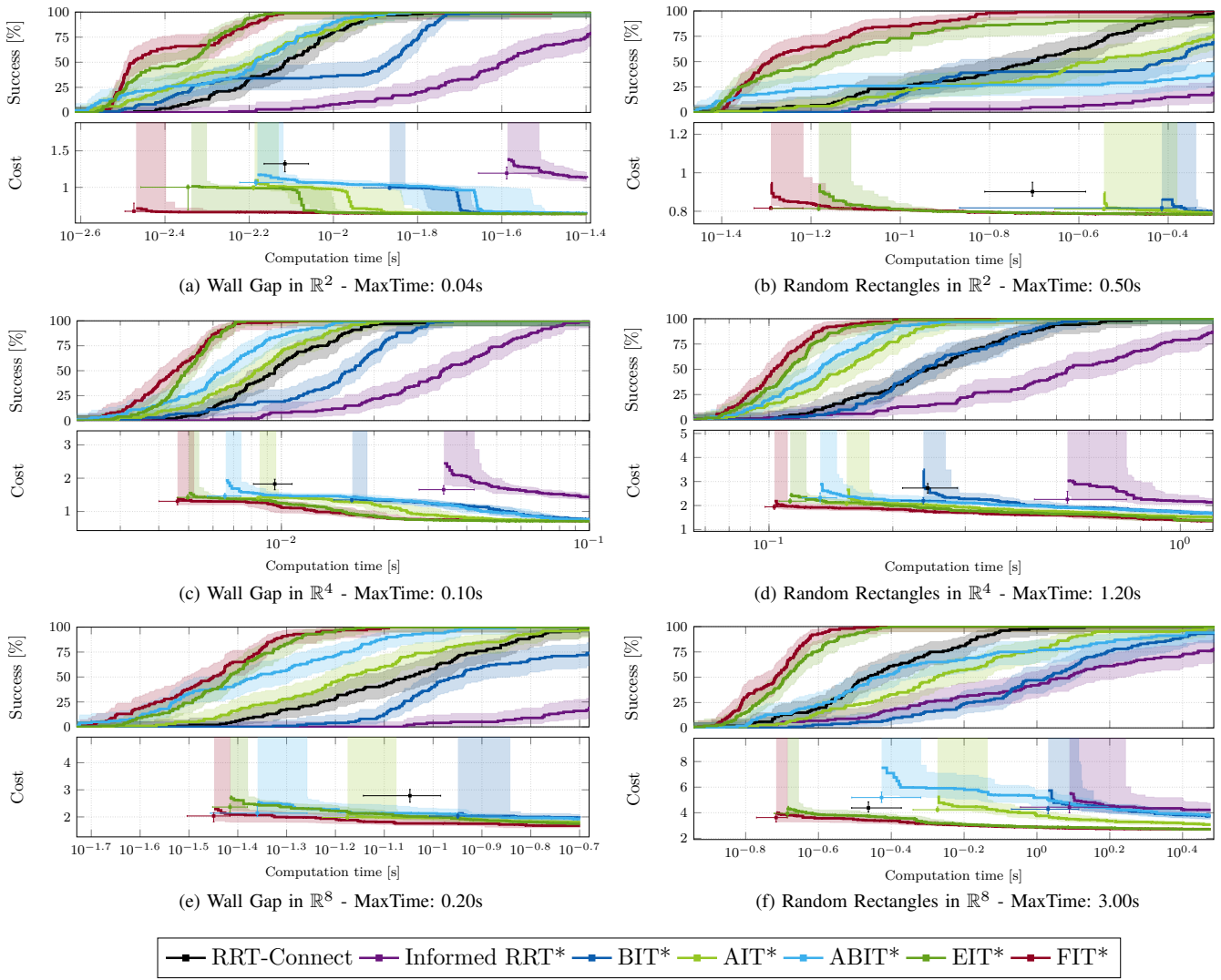
\begin{figure*}[t!]
    \centering
    \begin{tikzpicture}
    \node[inner sep=0pt] (russell) at (4.1,8)
    {\includegraphics[width=0.49\textwidth]{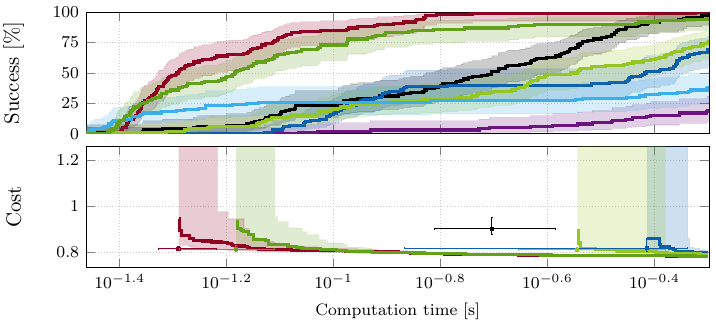}};
    \node[inner sep=0pt] (russell) at (4.1,3.5)
    {\includegraphics[width=0.49\textwidth]{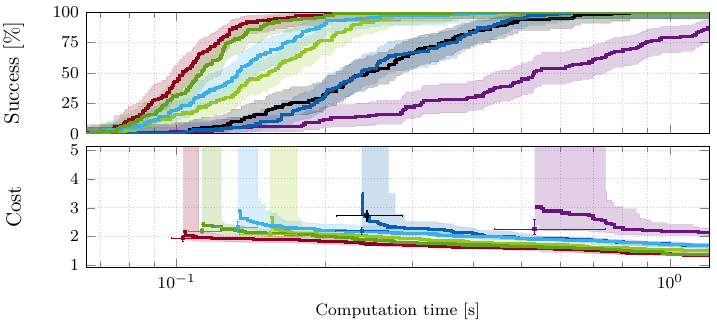}};
    \node[inner sep=0pt] (russell) at (4.1,-1)
    {\includegraphics[width=0.49\textwidth]{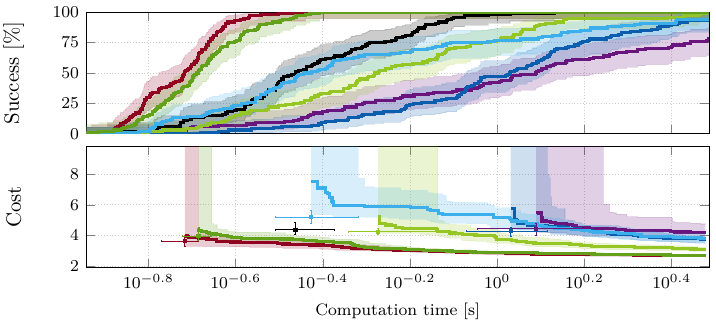}};

    \node[inner sep=0pt] (russell) at (-4.87,8)
    {\includegraphics[width=0.495\textwidth]{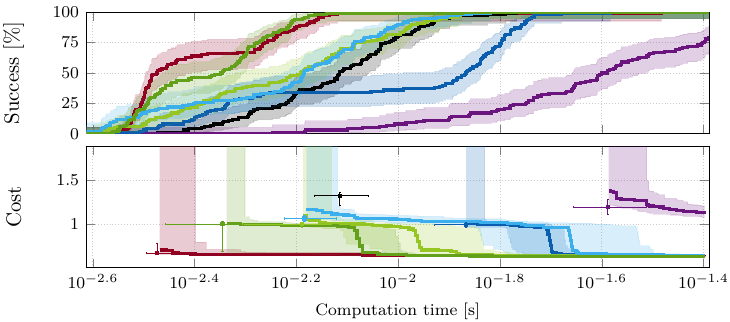}};
    \node[inner sep=0pt] (russell) at (-4.87,3.5)
    {\includegraphics[width=0.49\textwidth]{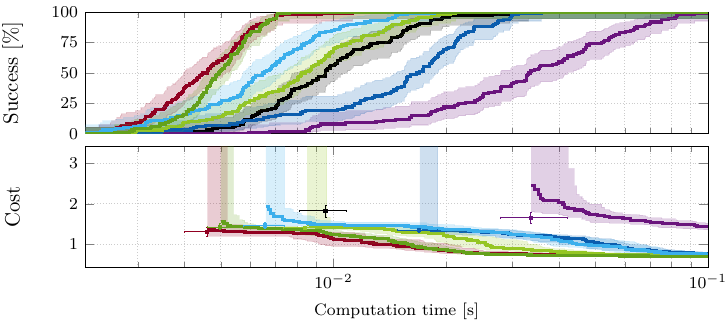}};  
    \node[inner sep=0pt] (russell) at (-4.9,-1){\includegraphics[width=0.49\textwidth]{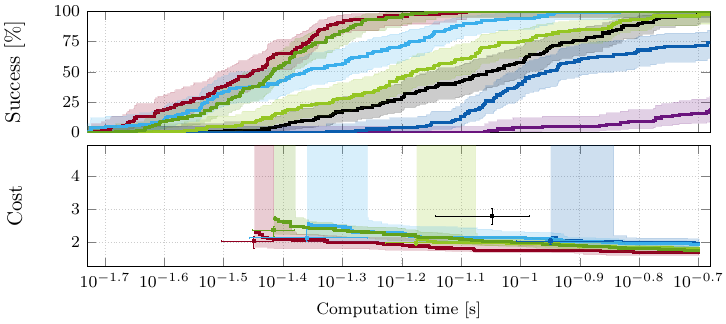}};

    \node[inner sep=0pt] (russell) at (0.0,-4.05){\includegraphics[width=0.8\textwidth]{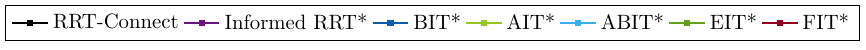}};

    \node at (-4.5,5.8) {\footnotesize (a) Wall Gap in $\mathbb{R}^2$ - MaxTime: 0.04s};
    \node at (-4.5,1.3) {\footnotesize(c) Wall Gap in $\mathbb{R}^4$ - MaxTime: 0.10s};
    \node at (-4.5,-3.2) {\footnotesize(e) Wall Gap in $\mathbb{R}^8$ - MaxTime: 0.20s};

    \node at (4.5,5.8) {\footnotesize (b) Random Rectangles in $\mathbb{R}^2$ - MaxTime: 0.50s};
    \node at (4.5,1.3) {\footnotesize(d) Random Rectangles in $\mathbb{R}^4$ - MaxTime: 1.20s};
    \node at (4.5,-3.2) {\footnotesize(f) Random Rectangles in $\mathbb{R}^8$ - MaxTime: 3.00s};

    \end{tikzpicture}
    \vspace{-1.0em} 
    \caption{Detailed experimental results from Section~\ref{subsec:experi} are presented above. Fig. (a), (c) and (e) depict test benchmark wall gap outcomes in $\mathbb{R}^2$, $\mathbb{R}^4$ and $\mathbb{R}^8$, respectively. Panel (b) showcases ten random rectangle experiments in $\mathbb{R}^2$, while panels (d) and (f) demonstrate in $\mathbb{R}^4$ and $\mathbb{R}^8$. In the cost plots, boxes represent solution cost and time, with lines showing cost progression for an almost surely optimal planner (unsuccessful runs have infinite cost). Error bars provide nonparametric 99\% confidence intervals for solution cost and time.}
    \label{fig: result}
    \vspace{-1.8em}
\end{figure*}

    Natural logarithm smoothly diminishes the decay rate, averting sudden drops for stable model optimization. The decay factor $\Psi_\text{decay}$ is determined through the division operation involving the logarithmic function. This relationship emphasizes how the dimensions of the hyperellipsoid influence the exploration of the state space in a logarithmic manner.
    \begin{equation}\label{fuc:decay}
        \Psi_\text{decay} = \frac{\ln(1+\Lambda \times \mathcal{O}_\text{smooth})}{\ln{(1+\Lambda)}},
    \end{equation}

    The sigmoid smoothing technology of the raw ratio $\mathcal{O}_\text{smooth}$ ensures a gradual transition between batch sizes, mirroring how the hyperellipsoids’s hypervolume changes smoothly with varying semi-axes lengths. This smoothing function allows for a more continuous adjustment of the batch size.

    \begin{equation}\label{fuc:smooth}
        \mathcal{O}_\text{smooth} = \frac{1}{1+ e^ {-10 \times {(\xi_n} -0.5)}},
    \end{equation}
    where $e$ is Euler’s number $e = \sum_{i=0}^\infty \frac{1}{n!}$

    Decay tuning parameter $\Lambda_\text{tuning}$ is correlated on the problem’s dimensionality $n_\text{dimension}$ along with the $m_\text{min}$ and $m_\text{max}$ samples. This tuning parameter is a composite representation that makes the batch size related to \textit{$\mathcal{C}$-space} dimensional information and becomes more problem-specific during adaptation.

    \begin{equation}
        \Lambda_\text{tuning} = \frac{\mathcal{M}(\Psi_\text{max})+\mathcal{M}(\Psi_\text{min})}{n_\text{dimension}},
    \end{equation}

    The adjustment of batch size responds to the current raw ratio $\xi_n$, aligning with the optimization phase of the problem. This ratio is affected by the contraction (i.e., solution cost update) of the hypervolume of the current $n$-dimensional hyperellipsoids, this raw ratio is represented as:
    
    \begin{equation}
        \xi_n = \frac{\mathcal{V}_\text{current}}{\mathcal{V}_\text{initial}}.
    \end{equation}

Given the continuous improvement of the solution cost, the hypervolume shrinks accordingly, $\mathcal{V}_\text{current}$ consistently remains less than or equal to $\mathcal{V}_\text{initial}$. Consequently, the current raw ratio of the process $\xi_n$ resides within the interval $(0,1]$.
\section{Experimental Results}\label{sec:Expri}
FIT* was tested against existing algorithms in both simulated random scenarios (Fig.~\ref{fig: testEnv}) and real-world manipulation problems (Fig.~\ref{fig:simulation}). The comparison involved several versions of RRT-Connect, Informed RRT*, BIT*, AIT*, ABIT*, and EIT* sourced from the Open Motion Planning Library (OMPL)~\cite{sucan2012open}. The evaluations are implemented on a computer with an Intel i7 3.90 GHz processor and 32GB of LPDDR3 3200 MHz memory. These comparisons were carried out in simulated environments ranging from $\mathbb{R}^2$ to $\mathbb{R}^8$. The primary objective for the planners was to minimize path length. The RGG constant $\eta$ was uniformly set to 1.1, and the rewire factor was set to 1.001 for all planners.

In the case of RRT-based algorithms, a goal bias of 5\% was employed, and the maximum edge lengths were appropriately determined based on the dimensionality of the space. Meanwhile, BIT*, AIT*, ABIT*, and EIT* maintained a fixed sampling of 100 states per batch, regardless of the dimensionality of the state space. These planners also had graph pruning deactivated and utilized the Euclidean distance and effort as heuristic functions.
FIT*'s adaptive batch size technology dynamically adjusts the batch size, displaying a range from 1 to 199 batch sizes (Fig.~\ref{fig: decay_method}). The specific number of samples at anytime is determined by the planner's adaptive mechanisms, ensuring quantities that are optimized for each solution cost update and batch re-sampling process.

\subsection{Experimental Tasks}\label{subsec:experi}
The planners were subjected to testing across three distinct problem domains: $\mathbb{R}^2$, $\mathbb{R}^4$, and $\mathbb{R}^8$. In the first scenario, a constrained environment resembling a wall with a narrow gap was simulated, allowing valid paths in two general directions for non-intersecting solutions (Fig. \ref{fig: testEnv}a). The path planning objective is to optimize the initial and final path length rapidly. Each planner underwent 100 runs, and the computation time for each asymptotically optimal planner is demonstrated in the labels, with varying random seeds. The overall success rates and median path lengths for all planners are depicted in Fig.~\ref{fig: result}a, \ref{fig: result}c, and \ref{fig: result}e.
In the second test scenario, random widths were assigned to \textit{axis-aligned hyperrectangles}, generated arbitrarily within the \textit{$\mathcal{C}$-space} (Fig. \ref{fig: testEnv}b). Random rectangle problems were created for each dimension of the \textit{$\mathcal{C}$-space}, and each planner underwent 100 runs for every instance. Fig.~\ref{fig: result}b, \ref{fig: result}d, and \ref{fig: result}f illustrate the overall success rates and median path costs within the computation time for all the planners.

FIT* employs an adaptive batch size feature, enhancing convergence time in the initial pathfinding phase by up to 24\% and achieving faster solutions with lower initial costs.

\subsection{Path Planning for DARKO}\label{subsec:realExpri}
FIT* demonstrated its effective adaptive batch size techniques during a field test as part of the inventory management (Fig.~\ref{fig: darko_setup} and Fig.~\ref{fig:simulation}). DARKO is a mobile manipulation robotic platform (8-DoF) created by combining Robotnik base robot and Franka Emika Panda manipulator, addressing intralogistics challenges. 
The intricacy of the narrow space presents challenging planning problems, primarily due to the computationally expensive state evaluations required. 
All planners had 1.0 seconds to address this confined, limited space pull-out and place problem. Over 15 trials, FIT* was 100\% successful with a median solution cost of 19.1021. EIT* was also 100\% successful but had a median solution cost of 22.6917. AIT* was 86.7\% successful with a median solution cost of 26.4482, and ABIT* was 80\% successful with a median solution cost of 25.6341.
 In contrast to other planners that rely on occupied space, FIT* showcased efficiency by saving working space and completing the task with the shortest optimal path length. The detailed behavior of DARKO can be viewed in the accompanying video.
\section{Discussion \string& Conclusion}
This paper introduces FIT*, an adaptive batch size method planner correlated to the \textit{$\mathcal{C}$-space}'s dimensionality and the hypervolume of the $n$-dimensional hyperellipsoid. In the initial solution finding phase, increased samples per batch accelerate initial solution discovery (higher probability to sample in the key area), while in the optimal phase, fewer samples per batch reduce collision checking time, enhancing sampling efficiency across batches. 
%
The adaptability of FIT* was exemplified in a real-world scenario with the DARKO robot. FIT*'s adaptive strategies ensured rapid initial solutions. 
This illustrates FIT*'s practical applicability and industry problem-solving capacity.

In conclusion, FIT* employs a flexible approach by dynamically optimizing batch sizes based on the sigmoid-log function to leverage a decay factor related to the $n$-dimensional hyperellipsoid. The adaptive batch size method showcases its potential by modifying the number of samples per batch and batch update frequency to optimize the planner's initial and optimization phases. This feature positions FIT* as a promising solution in the field of path planning.








\bibliographystyle{IEEEtran}
\bibliography{references}
\balance

\end{document}